\documentclass[10pt,twocolumn,letterpaper]{article}

\usepackage{iccv}              % To produce the CAMERA-READY version
\definecolor{iccvblue}{rgb}{0.21,0.49,0.74}
\usepackage[pagebackref,breaklinks,colorlinks,allcolors=iccvblue]{hyperref}

\usepackage{caption}
\usepackage{graphicx}
\usepackage{hhline}
\usepackage{multirow}
\usepackage{pifont}
\usepackage{xcolor}
\usepackage[normalem]{ulem}

\usepackage{cuted}
\usepackage{capt-of}

\usepackage{colortbl}

\def\paperID{9049} % *** Enter the Paper ID here
\def\confName{ICCV}
\def\confYear{2025}

\newcommand{\approach}{TOGA }

\title{TOGA: Temporally Grounded Open-Ended Video QA with Weak Supervision}

\author{
Ayush Gupta$^{1,2}$, 
Anirban Roy$^{1}$, 
Rama Chellappa$^{2}$, 
Nathaniel D. Bastian$^{3}$, 
Alvaro Velasquez$^{4}$, 
Susmit Jha$^{1}$\\
$^1$SRI \quad
$^2$Johns Hopkins University \quad
$^3$United States Military Academy \quad
$^4$University of Colorado Boulder \\
}

\begin{document}

\maketitle
\makeatletter
\renewcommand{\@makefnmark}{}
\makeatother
\footnotetext{Work partly done during Ayush Gupta's internship at SRI.}

%\begin{figure}
\begin{strip}\centering
\includegraphics[width=0.90\textwidth]{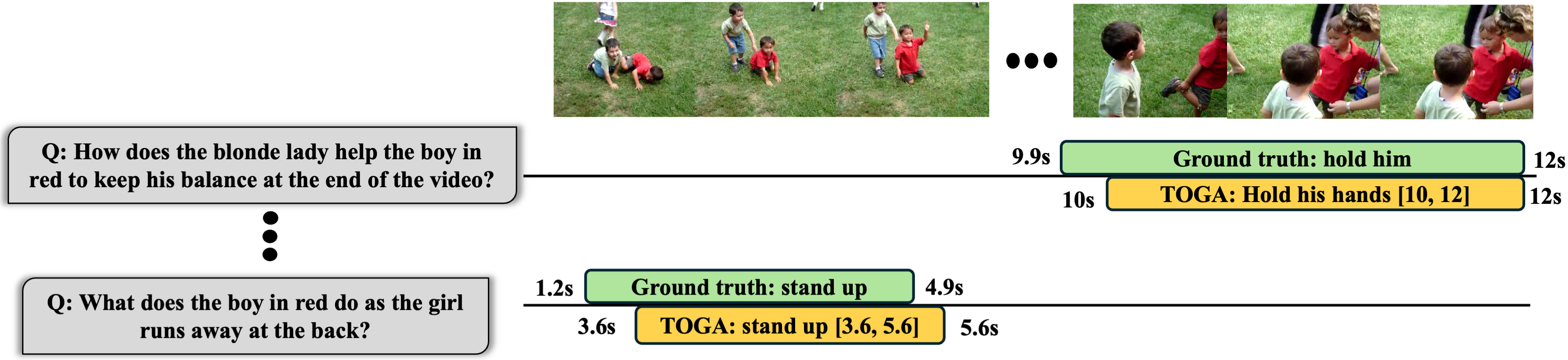}
\captionof{figure}{We present an approach for open-ended grounded video question answering. Given a video and open-ended questions, we generate open-ended responses with the grounding as \texttt{Answer [start time, end time]}. We consider long videos with multiple questions and answers per video. The questions refer to the interaction between multiple actors and the temporal ordering of events. 
%We consider a weakly supervised setup where grounding annotations are unavailable during training.
}
\label{fig:overview}
\vspace{-2mm}
\end{strip}
%\end{figure}

% \begin{figure*}[h]
%     \centering
%     \resizebox{\linewidth}{!}{
%         \includegraphics{figures/Teaser figure.png}
%     }
%     \caption{Caption}
%     \label{fig:teaser-fig}
% \end{figure*}

\begin{abstract}

We address the problem of video question answering (video QA) with temporal grounding in a weakly supervised setup, without any temporal annotations. Given a video and a question, we generate an open-ended answer grounded with the start and end time. 
For this task, we propose \approach: a vision-language model for Temporally Grounded Open-Ended Video QA with Weak Supervision. 
We instruct-tune \approach to jointly generate the answer and the temporal grounding. 
We operate in a weakly supervised setup where the temporal grounding annotations are not available.
We generate pseudo labels for temporal grounding and ensure the validity of these labels by imposing a consistency constraint between the question of a grounding response and the response generated by a question referring to the same temporal segment. 
We notice that jointly generating the answers with the grounding improves performance on question answering as well as grounding.
We evaluate \approach on grounded QA and open-ended QA tasks. 
For grounded QA, we consider the NExT-GQA benchmark which is designed to evaluate weakly supervised grounded question answering.
For open-ended QA, we consider the MSVD-QA and ActivityNet-QA benchmarks. We achieve state-of-the-art performance for both tasks on these benchmarks. 
\end{abstract}    
\section{Introduction}
\label{sec:intro}

We propose a framework for grounded video question answering (videoQA) under weak supervision. Given a video and a question as inputs, we generate an open-ended answer temporally grounded with the start and end times as in \cref{fig:overview}. Our answers are free-form sentences, not restricted to a single word/phrase or an option from multiple choices. To avoid reliance on expensive temporal annotations, we consider a weakly supervised setup without ground truth grounding annotations. 

% Why is the task challenging?
% 1.answer and localize
% 2.weakly supervised
% 3.Multiple temporal windows in same video
% 4.Open ended

Grounded video question answering is challenging as, in addition to generating correct answers, it requires localizing evidence to support the answer~\cite{nextgqa-dataset}. 
The challenge is more prominent in a weakly supervised setup where grounding annotations are unavailable. 
Since each video can have multiple questions with overlapping temporal groundings, each answer needs to be grounded with a distinct temporal window. We consider relatively long videos with an average length of 40 seconds consisting of complex causal and temporal questions~\cite{nextgqa-dataset}. 
Generating correct answers to these questions requires understanding the temporal ordering of events and the spatio-temporal interaction between actors and objects. 
Further, we consider open-ended evaluation instead of a multiple-choice QA setup like previous works~\cite{sevila,llovi}, further enhancing the challenge.

% How our approach solves the challenges
% 1.answer and localize - Instruction tuning format
% 2.weakly supervised - consistency loss
% 3.Multiple temporal windows in same video - multi-scale VLC
% 4.Open ended - Leverage large language decoder (LLM)

We propose \approach - a vision-language model (VLM) for the grounded videoQA task to address these challenges. \approach builds a video processing framework and combines that with a large language model (LLM)-based text processing framework to process questions and generate open-ended answers. 
We instruction-tune \approach to jointly generate answers with temporal groundings in the format: \mbox{\texttt{Answer [start time, end time]}}. 
Given a video and a question, we sample video frames and compute visual features using a pre-trained vision transformer encoder~\cite{CLIP}. The question is processed with the LLM tokenizer and embedding layer ~\cite{mistral-7b} to capture text features. 
Inspired by previous work \cite{feichtenhofer2019slowfast,xiao2020audiovisual, kazakos2021slow, lita}, we utilize a multi-scale vision-language connector (MS-VLC) to align the video and text features. MS-VLC processes the video at two granularities: one at a low frame rate to capture low-frequency temporal features and another at a high frame rate to capture high-frequency temporal features. 
Finally, we instruction-tune the LLM decoder~\cite{mistral-7b} to jointly generate open-ended answers with temporal grounding by leveraging the cross-attention between text features and the multi-scale video features. 
% MS-VLC is more effective at grounding answers with diverse temporal durations than processing the video at a single frame rate, as shown for other temporal tasks as well~\cite{feichtenhofer2019slowfast,xiao2020audiovisual, kazakos2021slow}. 
Jointly generating answers with groundings allows capturing the dependency between an answer and the corresponding grounding duration, improving QA and grounding accuracy compared to the approaches that independently predict answers and their grounding~\cite{frozenbilm,sevila,llovi}. 

%weakly supervised
To operate in a weakly supervised setup without grounding labels, we propose a multi-stage training approach.
Firstly, we train \approach to generate answers without grounding by leveraging the question-answer annotations and video descriptions. 
Then, this model is used to generate pseudo-labels for temporal grounding. 
We select temporal segments with specific starting and ending times and ask questions corresponding to each segment. The answers along with the selected starting and ending times are considered noisy grounding labels for the corresponding questions. 
Next, we instruction-tune the model to accept prompts with temporal references such as \texttt{What is the activity in [10, 20]?} and produce responses with temporal grounding \texttt{A boy in a red shirt is running [10, 20]}.
Instruction tuning extends the model's abilities to accept temporal references in questions and produce grounding predictions in the answers. 
However, the grounding performance is limited due to noisy temporal labels. 
We improve the grounding performance by imposing a consistency constraint while generating the pseudo labels. We select the labels where the answer with a temporal grounding matches a question with the same temporal segment as the reference. For example, let's assume the temporally grounded answer to the question \texttt{What is the boy in a red shirt doing?} is \texttt{The boy is running [10, 20]}. Then the answer to a corresponding referring question \texttt{What is happening in [10, 20]?} is expected to match with \texttt{A boy in a red shirt is running}.
Maintaining consistency between the answers to the referring and grounding questions is crucial in a weakly supervised setup as shown in our ablations studies. 
%\AG{Maybe refer to the multi-stage figure here}

%What current methods are lacking
\approach has several advantages over existing approaches. We jointly generate answers with the groundings. Thus, the model can adjust the temporal duration based on the answers by capturing the correlation between the answer and the grounding. Approaches making independent predictions~\cite{frozenbilm,sevila,llovi} may lack this ability.
We instruction-tune the language decoder to generate open-ended answers. Thus, we are not limited to restrictive formats of answers, such as choosing an answer from multiple choices or answering with single-word responses.
Finally, we use consistency constraints to generate pseudo-labels for training with weak supervision. \approach does not rely on temporal annotations or external models to generate annotations~\cite{groundedvideollm}.
% \input{ICCV/tables/capabilities-other-methods}

%We compare the task setup of \approach with the closely related vision-language models against various aspects of grounded video QA in \cref{tab:capabilities}.

Our main contributions include:

\begin{itemize}[leftmargin=*]%,topsep=0pt, itemsep=0pt]

\item We propose TOGA, a large vision-language model for open-ended grounded videoQA. \approach jointly generates open-ended answers with temporal grounding.
% trained by imposing a consistency constraint between QA and grounding responses.  
% \AG{Consistency? Mention that we perform weakly supervised grounding through our multi-stage training process?}

\item \approach operates in a weakly supervised setup where groundings annotations are unavailable. We train the model with reliable pseudo labels by  imposing consistency between the answers to temporal grounding and temporal referring questions.

\item We evaluate our approach on weakly supervised temporal grounding and video QA datasets, and advance the state-of-the-art results on these benchmarks.

%\item We generate open-ended answers that differ from existing approaches which select answers from a pre-defined set of options.

\end{itemize}

% % Overview
% % Recently, there have been a lot of advancements in LLMs \cite{gpt4, gemini, llama}. While they traditionally just operated on text input, there have been advances connecting them to 
% As video content becomes ubiquitous, it has become increasingly important to develop models which can understand visual input and connect it to natural language. There has been a lot of recent work on connecting image and language domains \cite{llava, ferret}, but these works fail to operate on video input with the added temporal dimension. 
% As a result, there has been more research in understanding videos, with several works \cite{videollava, videollama} providing solutions for this task.

% To generate a temporally grounded answer, we develop a Grounded Instruction Tuning (GIT) framework, where we train the model to generate timestamps in a predefined format.   % We train the model to do referring as well
% To train our approach in a weakly supervised manner, we propose to leverage the consistent answers in the QA and referring task \AG{Refine. We suddenly brought up referring without mentioning it before.}
% We utilize the MS-VLC module to look at multiple temporal scales and avoid confusion with the temporal windows of other questions. % Does this sound right? How does MSVLC help avoid confusion?
% Lastly, we leverage large-scale pretrained LLMs as our language decoder to generate open ended answers.
% \AG{Refine the above paragraph}
\section{Related Work}
\label{sec:related_work}
%\AG{TODO}
% VideoQA methods are predominantly
% banked on transformer and pre-training. The popular transformer architectures follow either shared 
% dual or stacked implementations, and pre-training is done with image-text, videotext or both forms of data. Notably, all these VLMs use powerful language models (e.g., BERT,
% T5, GPT, LLaMA or their successors) for text
% encoding and focus on improving QA while ignoring visual
% evidence grounding. Some recent works 
% have begun to ground key frames or objects for VideoQA.
% Yet, they still aim to improve QA accuracy, and thus the
% grounded contents may not be the actual evidences since they
% do not evaluate grounding. For weakly-supervised video
% grounding, typical approaches extract temporal proposals
% and rank the proposals according to their similarities with the
% language query. Despite their effectiveness,
% these two-stage approaches are notorious for inefficient and
% sub-optimal for multi-granular temporal modelling. More recent research points to the superiority of end-to-end
% Gaussian mask learning. In light of this, we design a simple
% yet effective Gaussian mask learning module for grounding in VideoQA. Unlike previous works  that design
% small transformer models and hand-craft negative visual proposals for contrastive learning, we integrate Gaussian mask
% learning into large VLMs and optimize its parameters via
% question-answering and video-question grounding

\begin{figure*}[t]
  \begin{center}
  \includegraphics[width=1\linewidth]{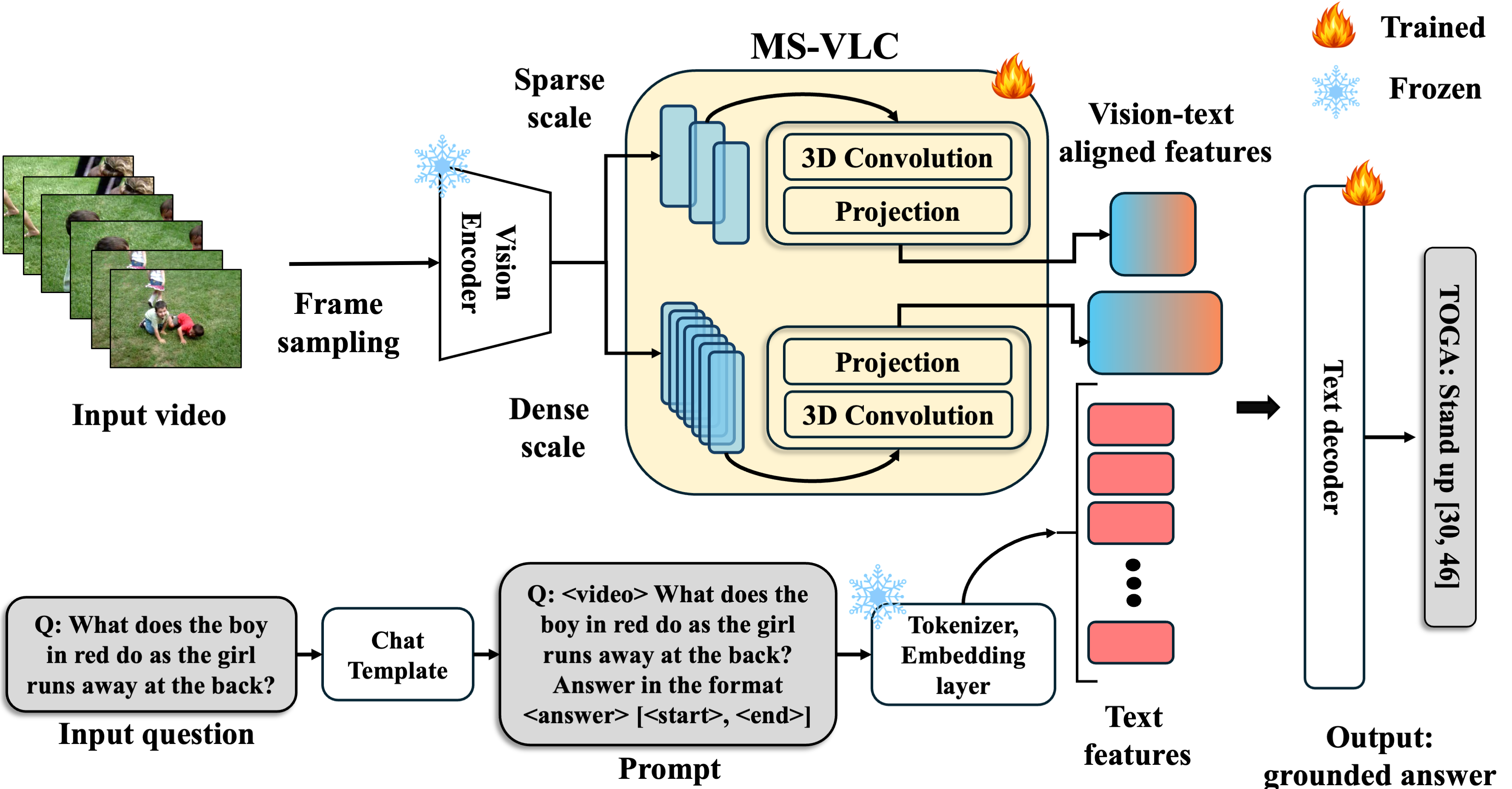}
  \end{center}
  \caption{
  We propose a VLM framework for grounded videoQA. Given an input video, we sample frames and compute framewise features using a vision encoder. A multi-scale vision language connector (MS-VLC) is employed to process the framewise features at two temporal resolutions: a sparse and a dense scale. The input question is processed with a chat template according to the desired prompt format. The prompt is processed through a tokenizer and a frozen language encoder to generate tokenwise text features. The MS-VLC module is trained to align the multiscale vision and text features. Finally, the language decoder is trained to generate answers with temporal grounding.
  %\AG{Add sys prompt in prompt box}
  %\AG{Use same video as teaser figure, and add VIGOR to the output. Label the sparse connector and dense connector. Remove whitespace}
  \vspace{-2mm}
  }
  \label{fig:framework}
\end{figure*}

\textbf{Video question answering.}
%include Video LMs in this section
Video QA aims to answer questions related to the visual content in videos. Compared to image visual question answering \cite{vqa-1, vqa-2}, videoQA requires capturing the temporal evolution of scenes. Several approaches are proposed to address this task \cite{tgif-qa, video-vqa-2, nextqa, ovqa-benchmark, videoqa-survey, mmvqa-1, tvqa, tvqa-plus, kbvqa-1, kbvqa-2}. 
Some approaches  only utilize the visual cues for answering questions \cite{tgif-qa, video-vqa-2, nextqa}. 
However, others utilize additional modalities, like transcripts or subtitles of videos, or the movie plot \cite{mmvqa-1, tvqa, tvqa-plus}. 
Utilizing external knowledge bases~\cite{wordnet, conceptnet} is shown to be effective in improving the answering performance~\cite{kbvqa-1, kbvqa-2}. However, the videoQA approaches commonly consider a multiple-choice setup, where the task is to select a candidate answer from a set of predefined options. We consider an open-ended setup and generate free-form answers.

\textbf{Open-ended video QA}
More recently, with the advancement of LLMs \cite{gpt4, gemini, llama}, videoQA approaches aim to generate open-ended answers~\cite{videollava, videollama2, videochatgpt, videochat, frozenbilm, videollama, vipergpt, videoagent, videotree, videoqa-llm-survey}.
These methods combine vision module and language models to develop general-purpose VLMs~\cite{videollava, videollama2, videochatgpt, videochat, videollama}. These models are typically trained on large-scale datasets with various multimodal tasks. 
Yang et al.,~\cite{frozenbilm} develop specialized models that are fine-tuned for the target video QA task.
There are parameter-free methods that use the collaboration between multiple VLM agents \cite{vipergpt, videotree, videoagent} for the video QA task. However, many current approaches are not able to provide evidence for the generated answers in the video in the form of grounding and may be relying on language priors instead of true visual reasoning \cite{videoqa-llm-survey}. 
We focus on addressing this issue by grounding the answers in videos with temporal segments.

\textbf{Grounded video QA.}
Grounded video QA aims to generate answers and ground them in the video with temporal segments. Given the question, the segment is expected to contain the evidence for the answer.
Recent approaches aim to address this by exploring the similarity between visual and textual content~\cite{tvqa, tvqa-plus, vidstg}. However, these approaches tend to be biased towards localizing subtitles in TV shows \cite{tvqa} or only being able to deal with a few objects \cite{vidstg}. 
Recently, some works have been proposed which leverage the power of VLMs for the grounded video QA task \cite{lita, numberit, groundedvideollm, videostreaming}.
However, these approaches require expensive temporal annotations for training, or need to generate grounding annotations using external methods.
To overcome this, some works incline towards weakly supervised setup where temporal annotations are not available~\cite{nextgqa-dataset, sevila, llovi}. However, these approaches consider a multiple-choice setup, where the answer choices are provided to the model during evaluation.

Unlike existing approaches, \approach aims to generate temporally grounded, open-ended answers in a weakly supervised setup without ground truth temporal annotations.
%To the best of our knowledge, no other work addresses this task.
%\AG{Shorten}

\section{Approach}
\label{sec:approach}

\begin{figure*}[t]
    \centering
    \includegraphics[width=0.96\linewidth]{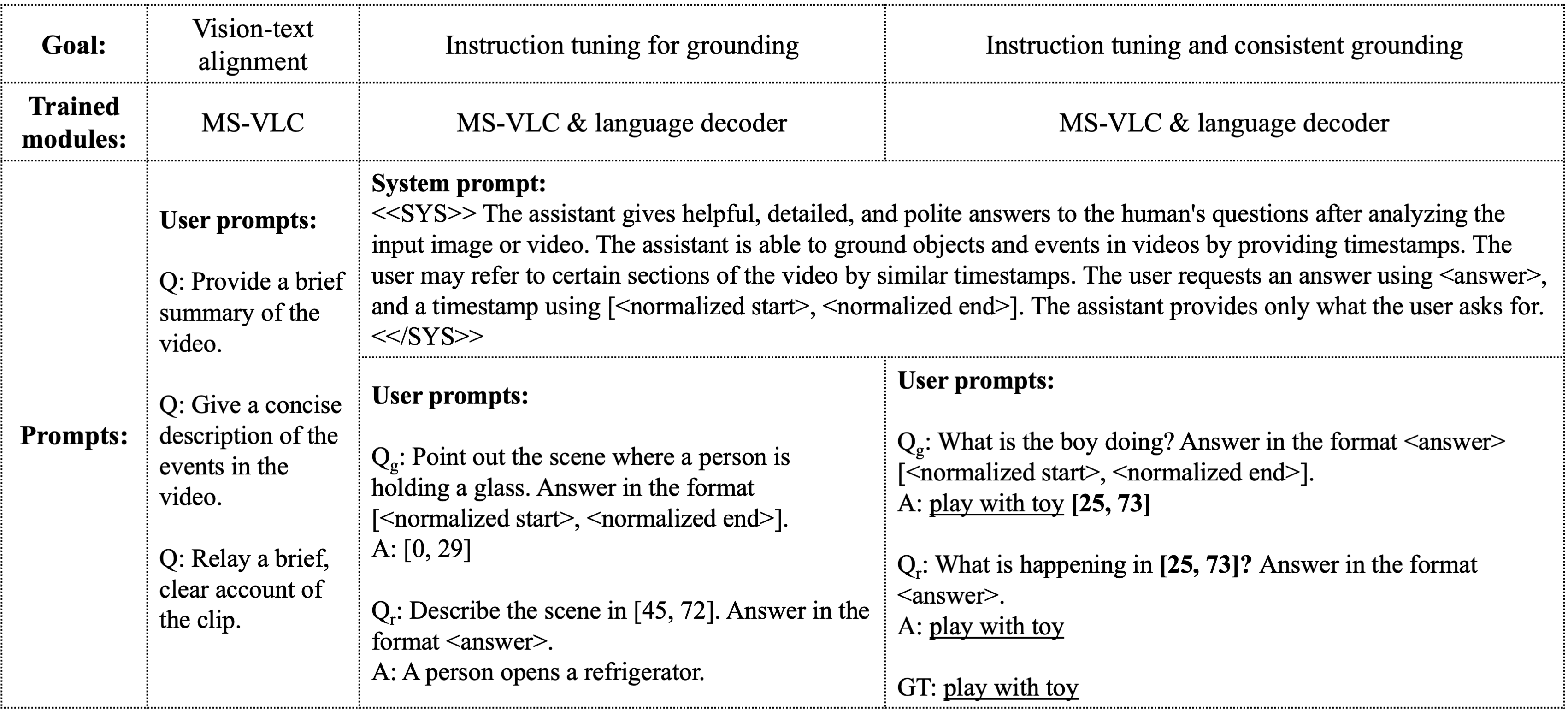}
    \caption{We train \approach in three stages. Each stage focuses on a specific task with prompts designed for the
    task. 
    Q\textsubscript{g} is a grounding question where a temporal grounding is expected in the response, such as \texttt{What is the boy doing?}. Q\textsubscript{r} is a referring question focusing on a temporal segment in the video, such as \texttt{What is happening in [25, 73]}.
    We impose consistency on the answers to Q\textsubscript{g} and Q\textsubscript{r}. These two queries are expected to have the same answer \texttt{\underline{play with toy}}. Further, we ensure the answer matches the ground truth (GT) answer. The multistage training gradually enables question answering, temporal referring, and temporal grounding. Temporal referring and temporal grounding enable grounding with weak supervision.  %\AG{Big, boring figure. Improve?}
    \vspace{-2mm}
    }
    \label{fig:multi-stage}
\end{figure*}

%This section presents the VLM framework, training strategy, and prompt design. 
Our VLM framework has four main modules: 1) a vision encoder to compute frame-wise features from videos, 2) a large-language text encoder to compute text features from questions, 3) a multi-scale vision-language connector (MS-VLC) to align the vision and text features facilitating answer generation, and 4) a large-language text decoder to generate open-ended answers with grounding as \mbox{\texttt{Answer [start time, end time]}}. 
Our approach is trained using our proposed multi-stage training framework leveraging consistency between predictions to learn the grounding ability without temporal labels.
Among the four components, vision and text encoders are kept frozen while the MS-VLC and text decoder are trained, like previous works \cite{llava}. The framework is shown in \cref{fig:framework}. We describe the components below.

\textbf{Vision encoder.} 
Given a video, we uniformly sample frames and compute frame-wise features. We choose the CLIP~\cite{CLIP} vision encoder that is trained with large-scale vision-language datasets. We keep this encoder frozen as it already generates the text-aligned features. 

% This gives us the flexibility of sampling these frames at arbitrary rates, which is essential for our multi-scale vision language connector as described later in this section.

\textbf{Text encoder.} 
We process the question with a LLM text encoder. 
The text encoder combines the tokenizer and the embedding layer of an LLM. This module generates text features that feed to the text decoder. Following the common practice~\cite{llava, videollama2, videollava}, we keep this module frozen as it is pre-trained on large and diverse text datasets.

% \AG{The `text encoder' is usually seen as part of the LLM, called the embedding layer. Having this as a separate module might confuse the reader, who may think we are doing something different.}

\textbf{Multi-scale vision-language connector (MS-VLC).}
Given the framewise features, we aim to capture the temporal cues to generate answers with temporal grounding. This module is trained to generate video features that are aligned with the text features derived from the question.  
The MS-VLC module processes the framewise features at two temporal resolutions: a sparse scale with a low frame rate to capture low-frequency temporal cues and a dense scale with a high frame rate to capture high-frequency temporal cues. We sample 4 frames at the sparse scale and 16 frames at the dense scale. The sparse module captures the long-term temporal features suitable for grounding longer segments, while the dense module captures the short-term temporal features suitable for grounding shorter segments. Each VLC block is implemented with RegNet~\cite {regnet} and 3D convolutions~\cite{videollama2}. The parameters are shared between the two VLC blocks. Our experiments show that the MS-VLC is crucial for both accurate answering and grounding. Processing time series at multiple scales is shown to be effective for  activity recognition in videos~\cite{feichtenhofer2019slowfast,xiao2020audiovisual} and event detection in audio~\cite{kazakos2021slow}.

%We set the dense sampling rate to be 4X higher than the sparse rate. 

% a 3D convolution with RegNet connections followed by a multi-layer perceptron (MLP) in this module. To capture multiscale cues, we sample framewise features at two rates: 1X and 2X. Sampled framewise features are processed by two MS-VLC blocks with shared weights. The outputs are high-frequency and low-frequency temporal features.

\textbf{Text decoder.} We train an LLM text decoder to generate answers from multi-scale video features and token-wise text features. We consider Mistral-7B Instruct \cite{mistral-7b} as the language decoder. 
This module learns the cross-attention between the multi-scale video features and the text features to generate answers in the desired format. 
This allows us to jointly generate open-ended answers with temporal grounding, unlike the approaches with frozen text decoder~\cite{frozenbilm,sevila,llovi} that generate the answer and grounding separately. We train the model on the next token prediction task as commonly used in language models \cite{causal-lm,llama,gpt4}.

\textbf{Multi-stage training.} 
We propose a multi-stage training strategy for temporally grounded Video QA under weak supervision.
\textbf{First}, we train only the MS-VLC module to align multiscale video and text features similar to \cite{llava}. This alignment is crucial for the downstream tasks. Given a question input, the goal is to generate aligned visual features enabling the text decoder to produce corresponding answers. We consider diverse prompt and response pairs for training at this stage, which includes video captioning, sentence completion, and question answering. 
\textbf{Second}, we instruction-tune the MS-VLC and the language decoder modules for the grounding task. The main goal of this stage is to train the model to understand prompts with temporal references such as \mbox{\texttt{What does the boy do at [10, 20]?}} and produce responses with temporal grounding \mbox{\texttt{A boy in a red shirt is running [10, 20]}}. As we do not have temporal annotations in the weakly supervised setup, we generate pseudo temporal labels for training. We crop temporal segments in videos with known start and end times. We generate descriptions for these segments by considering them as full videos by using the model trained in the previous stage. These descriptions with the selected start and end times are considered as pseudo labels for answers and temporal grounding, respectively. 
\textbf{Finally,} we train for accurate grounding by imposing a consistency constraint between the grounding response and the response generated by a question with the same grounding as input. For example, let's consider a response \mbox{\texttt{Stands up [5, 10]}} corresponding to a query \texttt{What does the boy in red do after the girl left?} Then we generate a paired question as \; \mbox{\texttt{What does the boy do in [5, 10]?}} with the same start and end times. We train the model to produce a consistent response \mbox{\texttt{Stands up}}. Furthermore, based on the available question-answer annotations, we ensure the answer {\texttt{Stands up}} is accurate. These self-consistent question-answers with temporal labels enable \approach to improve both answering and grounding accuracy.

% \AG{Mention the loss function - Causal language modelling using Cross entropy}

\textbf{Prompt design.}
We design prompts suitable for the tasks at various stages of training as shown in \cref{fig:multi-stage}. 
We define a special \texttt{<video>} token to include visual features with text tokens, like~\cite{videollava, videollama2}. 
For vision-text alignment, we consider multiple user prompts to train the MS-VLC module. 
For instruction tuning, we design prompts to include temporal references. 
Since we ask the model to output the grounding as text tokens, defining a specific format for the grounding outputs is necessary. This format is provided as part of the prompt in the individual stages of training.
% We first design a system prompt to describe the task as shown in \cref{fig:multi-stage}. 
%We consider specific user prompts for each stage of training. 
Specifically, we include the text `Answer in the format \texttt{<format>}' as part of the user prompt, specifying the format based on the task. We notice that including the output format in the user prompt is important to generate grounding responses. We use the following formats: 
1)~\texttt{answer} when we only generate the answer, 2)~\texttt{[$<$start$>$, $<$end$>$]} when we only generate temporal grounding, or
3) \texttt{answer [$<$start$>$, $<$end$>$]} when we generate an answer with grounding.

%The first two types of prompts are used for the instruction tuning stage when the model is being trained to understand the concept of a timestamp. Prompt 3 is important during the consistency stage and for obtaining the grounded answer during inference.

\textbf{Inference.}
% \AG{Add details like temperature, max p, sampling etc}
We use the same system prompt during inference.
We add instructions specifying the format of the response as described in the previous section. For grounded videoQA, we prompt the model to perform both grounding and answering. For open-ended videoQA on MSVD-QA and ActivityNet-QA, we prompt the model to generate only answers. 
Inference takes 0.6 seconds on average to generate the grounded answers on an A100 GPU.

\textbf{Implementation details.}
We use \texttt{CLIP-ViT-Large} \cite{CLIP} as the vision encoder
and Mistral-7B Instruct~\cite{mistral-7b} as the language model. MS-VLC comprises two RegNet stages, separated by a 3D convolutional layer. We sample 16 frames at the dense scale and 4 frames at the sparse scale. Each frame undergoes preprocessing, including resizing and cropping to a size of 336x336. In the first vision-text alignment stage, we train for one epoch with a batch size of 256 using the AdamW optimizer~\cite{loshchilov2017decoupled} with a cosine learning rate scheduler. The batch size is reduced to 128 for the two following stages of instruction tuning for grounding. The learning rate is set at 1e-3 for the alignment stage and lowered to 2e-5 for the instruction tuning stages. We apply a warmup ratio of 0.03 in all stages. Following the common practice~\cite{videochatgpt, videollava}, we use video-text pairs from Video-ChatGPT~\cite{videochatgpt} to train MS-VLC at the first stage of training. This dataset contains captioning prompts and generic QA tasks without any temporal grounding. The first stage of training takes 54 hours, and the latter two stages take 7 hours each with 8 A100 GPUs. Additional implementation details are provided in the supplemental material.

\section{Experiments}
\label{sec:experiments}

We first describe the datasets and metrics, compare \approach with the state-of-the-art, and perform analysis and ablations studies. Then, we present failure cases and discuss limitations and future directions. 

\subsection{Datasets and metrics}
\label{sec:datasets}
We evaluate \approach on four videoQA benchmarks: 
1)~NExT-GQA~\cite{nextgqa-dataset} 
and
2)~RexTime~\cite{rextime} for grounded QA, 
3)~MSVD-QA~\cite{msvd-msrvttqa} and 
4)~ActivityNet-QA~\cite{activitynet-qa} for open-ended QA.

\textbf{NExT-GQA~\cite{nextgqa-dataset}.} 
This is designed to evaluate weakly supervised grounded videoQA. Unlike other videoQA benchmarks \cite{jang2017tgif,wu2024star,xu2017video} consisting of short 3-15 second-long videos, NExT-GQA selects long videos with an average length of 40 seconds. 
Each video has multiple questions, and the answers' grounding can overlap. The videos include a sequence of atomic events, and the questions involve interaction between multiple actors and objects. 
Questions are of two types: causal why/how questions and temporal questions with when/before/after clauses. Causal questions, such as \texttt{Why are there two men standing at the center island and holding their camera?}, require localizing the evidence of the answer, which is \texttt{Recording} for this question.
Temporal questions, such as \texttt{What did the boy do after the green man walked past him?}, require understanding the temporal evolution of the events to generate an answer such as \texttt{Look at man in green}. 
The train set consists of 3,870 videos and 34,132 QA pairs without the grounding annotations. The test set consists of 990 videos with  5,553 QA pairs. 
Though the dataset provides multiple-choice answers, we generate open-ended responses without observing the options.

\textbf{ReXTime~\cite{rextime}.}
ReXTime is another benchmark designed to evaluate AI models' temporal reasoning abilities within video events. It specifically focuses on the challenging scenarios where questions and corresponding answers occur in different video segments, necessitating an understanding of cause-and-effect relationships across time. 
The benchmark comprises 921 validation samples and 2,143 test samples.
We utilize this dataset for zero-shot evaluation, and directly evaluate our model on the test samples without any fine tuning on this dataset.

\begin{table*}
\centering
\resizebox{0.8\linewidth}{!}{%
\begin{tabular}{c|c|ccccc}
 & \textbf{Open-ended evaluation} & \textbf{mIoU} & \textbf{mIoP} & \textbf{IoU@0.5} & \textbf{IoP@0.5} & \textbf{Acc@GQA} \\ 
\hline
IGV \cite{igv} \textit{(MM '22)} & \ding{55} & 14 & 21.4 & 9.6 & 18.9 & 10.2 \\
Temp[CLIP](NG+) \cite{CLIP,nextgqa-dataset} \textit{(CVPR '24)} & \ding{55} & 12.1 & 25.7 & 8.9 & 25.5 & 16 \\
FrozenBiLM (NG+) \cite{frozenbilm,nextgqa-dataset} \textit{(CVPR '24)} & \ding{55} & 9.6 & 24.2 & 6.1 & 23.7 & 17.5 \\
SeViLA \cite{sevila} \textit{(NeurIPS '23)} & \ding{55} & 21.7 & 29.5 & 13.8 & 22.9 & 16.6 \\
LLoVi \cite{llovi} \textit{(arXiv '24)} & \ding{55} & 20 & 37.3 & 15.3 & 36.9 & 24.3 \\
Grounded-VideoLLM \cite{groundedvideollm} \textit{(arXiv '24)} & \ding{55} & 21.1 & 34.5 & 18 & 34.4 & \textbf{26.7} \\
VideoStreaming \cite{videostreaming} \textit{(NeurIPS '24)} & \ding{55} & 19.3 & 32.2 & 13.3 & 31 & 17.8 \\ 
\hline
\textbf{\approach (Ours)} & \textbf{\checkmark} & \textbf{24.4} & \textbf{40.5} & \textbf{21.1} & \textbf{40.6} & 24.6
\end{tabular}
}
\caption{\label{tab:nextgqa-main}
Comparison with the state of the art on NExT-GQA \cite{nextgqa-dataset}. \approach improves the state of the art on the grounding metrics. Other approaches select an answer from a fixed set of options, while \approach generates open-ended answers. 
Note that \cite{groundedvideollm} uses ground truth labels for temporal grounding from other datasets (e.g., ActivityNet-Captions \cite{activitynet-captions}) and creates grounding labels with GPT-4 guidance.
\cite{videostreaming} utilizes ground truth temporal labels from Panda-70M \cite{panda70m} to curate grounding labels.
\approach does not use explicit grounding labels.
}
\vspace{-3mm}
\end{table*}

\textbf{Metrics for grounded videoQA.} 
We consider five metrics: intersection over union (IoU), intersection over prediction (IoP), IoU@0.5, IoP@0.5, and Acc@GQA proposed in NExT-GQA~\cite{nextgqa-dataset}. 
For ReXTime~\cite{rextime}, we use IoU, IoU@0.3 and IoU@0.5.
IoU measures the overlap between ground truth and predicted groundings. 
IoP measures the portion of predicted grounding containing the ground truth, similar to precision. 
IoU@0.5 and IoP@0.5 refer to the cases where IoU and IoP $\geq 0.5$. While these metrics focus on grounding, Acc@GQA focuses on both the correctness of the answers and correct grounding with \mbox{IoP $\geq 0.5$}.
Apart from these metrics, we consider the Acc@QA metric for QA performance in the supplementary section.

\textbf{MSVD-QA~\cite{msvd-msrvttqa}.}
This dataset considers open-ended video QA where the videos are selected from the video-description pairs of the MSVD dataset \cite{msvd}. QA pairs for a video are generated from the associated descriptions~\cite{heilman2009question}. It includes 1,970 video clips and 50K+ QA pairs. 

\textbf{ActivityNet-QA~\cite{activitynet-qa}.}
This dataset considers open-ended video QA with videos selected from ActivityNet~\cite{activitynet}. The videos are collected from YouTube. The dataset consists of 5,800 annotated videos and 58K QA pairs. The QA pairs are crowd-sourced by human annotators. 

\textbf{Metrics for open-ended QA.}
We consider two metrics: accuracy and score. An LLM generates a yes/no response by comparing the ground truth and predicted answers. The percentage of `yes' responses is the accuracy metric. The LLM also provides a score between 1 to 5 for the comparisons. The average score is considered as the metric. 
 
\textbf{LLM-based evaluation.} 
Commonly, grounded videoQA approaches consider a closed-set setup where answer options are available during inference, and the goal is to choose the correct option~\cite{nextgqa-dataset,sevila,llovi}. 
To compare our open-ended approach with other closed-set methods, we need to choose an option among the available choices and calculate the metrics.
For this, we use another pretrained LLM to select an option with the highest similarity to our prediction.
% \AG{Mention that we only do this to compare with current grounded VQA approaches}
Such LLM-assisted evaluation is commonly used in open-ended videoQA~\cite{videollava, videollama2, videochatgpt}. We use GPT-3.5-turbo~\cite{gpt4} to be consistent with these. We also experiment with the openly available LLama 3.1~\cite{llama} for evaluation and present the results in the supplemental material.

% All approaches answer questions in NExT-GQA by choosing from a set of five options. The method proposed in OVQA\cite{ovqa-benchmark} performs a retrieval-based technique, which chooses the answer from the (albeit large) test set vocabulary.

% \AG{Check if you are including this para or not, depending on LLama3.1/GPT-open eval results.}
% We also perform the  GPT-assisted evaluation similar to what is done in \cite{videollava, videollama2}. Specifically, given the question, our prediction, and the ground truth answer, we ask GPT to output a binary `yes' or `no' decision to decide whether our prediction matches the meaning of the ground truth. We report the percentage of `yes' answers as the accuracy. 
% %\AG{Perform GPT evaluation, otherwise remove this line!}

\subsection{Comparison with state of the art}

\textbf{Grounded videoQA.} We compare \approach with the state of the art on NExT-GQA~\cite{nextgqa-dataset} and show the result in \cref{tab:nextgqa-main}. \approach outperforms the existing approaches on grounding and on generating correct answers. 
 LLoVi \cite{llovi} is a zero-shot approach, while other methods are weakly supervised. 
It should be noted that our open-ended setup is more challenging than the closed-set setup followed in the existing approaches since the model is not able to view the options while generating the answer.
Additionally, other approaches generate the grounding output separately - either using a post-hoc approach, or using separate modules for answering and grounding. However, we generate the answer and the grounding jointly.
% \AG{Other approaches generate grounding in post-hoc/ad hoc manner separately.}
We believe that capturing the multi-scale temporal features and instruction-tuning the model to jointly predict answers with the grounding helps us achieve this performance.

\begin{table}
\centering
\resizebox{\linewidth}{!}{%
\begin{tabular}{c|c|ccc}
\multicolumn{1}{l|}{} & Model & mIoU & R@1 (IoU=0.3) & R@1 (IoU=0.5) \\ 
\hline
\multirow{2}{*}{Non Generative} & UniVTG & \uline{28.17} & \uline{41.34} & \uline{26.88} \\
 & CG-DETR & 23.87 & 31.31 & 16.67 \\ 
\hline
\multirow{4}{*}{LLM based} & VTimeLLM & 20.14 & 28.84 & 17.41 \\
 & TimeChat & 11.65 & 14.42 & 7.61 \\
 & LITA & 21.49 & 29.49 & 16.29 \\ 
\cline{2-5}
 & \textbf{Ours} & \textbf{25.53} & \textbf{29.91} & \textbf{19.79}
\end{tabular}
}
\caption{\label{tab:rextime}
Comparison with the state of the art on ReXTime \cite{rextime}, on zero-shot grounding. We compare our method with other generative LLM based grounding methods. 
The best generative scores are shown in bold.
We also include non-generative vision language grounding models for reference. The best non-generative model scores are underlined.
}
\end{table}

\textbf{Zero-shot grounding.}
We evaluate our approach on the ReXTime benchmark \cite{rextime} on the task of zero-shot query grounding. 
We compare our method with other generative LLM based methods in \cref{tab:rextime}. 
We also include non-generative grounding models for reference, but it is important to note that these non-generative methods can not answer open-ended queries like the generative methods.

%\AG{Talk about other approaches - they are zeroshot/weakly supervised?}

\textbf{Open ended videoQA.}
We consider MSVD-QA and ActivityNet-QA for this task. As shown in \cref{tab:open-ended-vqa}, \approach outperforms the state of the art on both datasets. We believe that capturing the multi-scale features with the MS-VLC enables \approach to accurately answer questions. 
%\AG{Ablation without MS-VLC? This gives lower score in QA.}

\begin{table}
\centering
\resizebox{\linewidth}{!}{%
\begin{tabular}{c|cc|cc}
\multirow{2}{*}{Method} & \multicolumn{2}{c|}{MSVD-QA} & \multicolumn{2}{c}{ActivityNet-QA} \\
 & Accuracy & Score & Accuracy & Score \\ 
\hline
FrozenBiLM \cite{frozenbilm} \textit{NeurIPS '22} & 32.2 & - & 24.7 & - \\
VideoChat \cite{videochat} \textit{(arXiv '23)} & 56.3 & 2.8 & - & 2.2 \\
LLaMA-Adapter \cite{llama-adapter} \textit{(ICLR '24)} & 54.9 & 3.1 & 34.2 & 2.7 \\
Video-LLaMA \cite{videollama} \textit{(EMNLP '23)} & 51.6 & 2.5 & 12.4 & 1.1 \\
Video-ChatGPT \cite{videochatgpt} \textit{(ACL '24)} & 64.9 & 3.3 & 35.2 & 2.7 \\
Chat-UniVi \cite{chat-univi} \textit{(CVPR '24)} & 65 & 3.6 & 45.8 & 3.2 \\
Video-LLaVA \cite{videollava} (\textit{EMNLP '24)} & 70.7 & 3.9 & 45.3 & 3.3 \\
Video-LLaMA2 \cite{videollama2} \textit{(arXiv '24)} & 70.9 & 3.8 & 50.2 & 3.3 \\ 
\hline
\textbf{Ours} & \textbf{73.8} & \textbf{3.9} & \textbf{52.0} & \textbf{3.4}
\end{tabular}
}
\caption{\label{tab:open-ended-vqa}
Comparison with the state of the art for open-ended videoQA on MSVD-QA and ActivityNet-QA. We outperform existing approaches against both metrics.
}
\vspace{-3mm}
\end{table}

% \AG{Mention that other approaches take options as input - we generate free form, then ask GPT to evaluate correct answer. So our evaluation is tougher, but still we are better.}

% \subsection{Open-ended video question answering.}
% %Section for GPT-assisted open eval/llama eval results
% \AG{Unsure if this section will be included}

\begin{figure*}
    \centering
    \resizebox{0.93\linewidth}{!}{
        \includegraphics{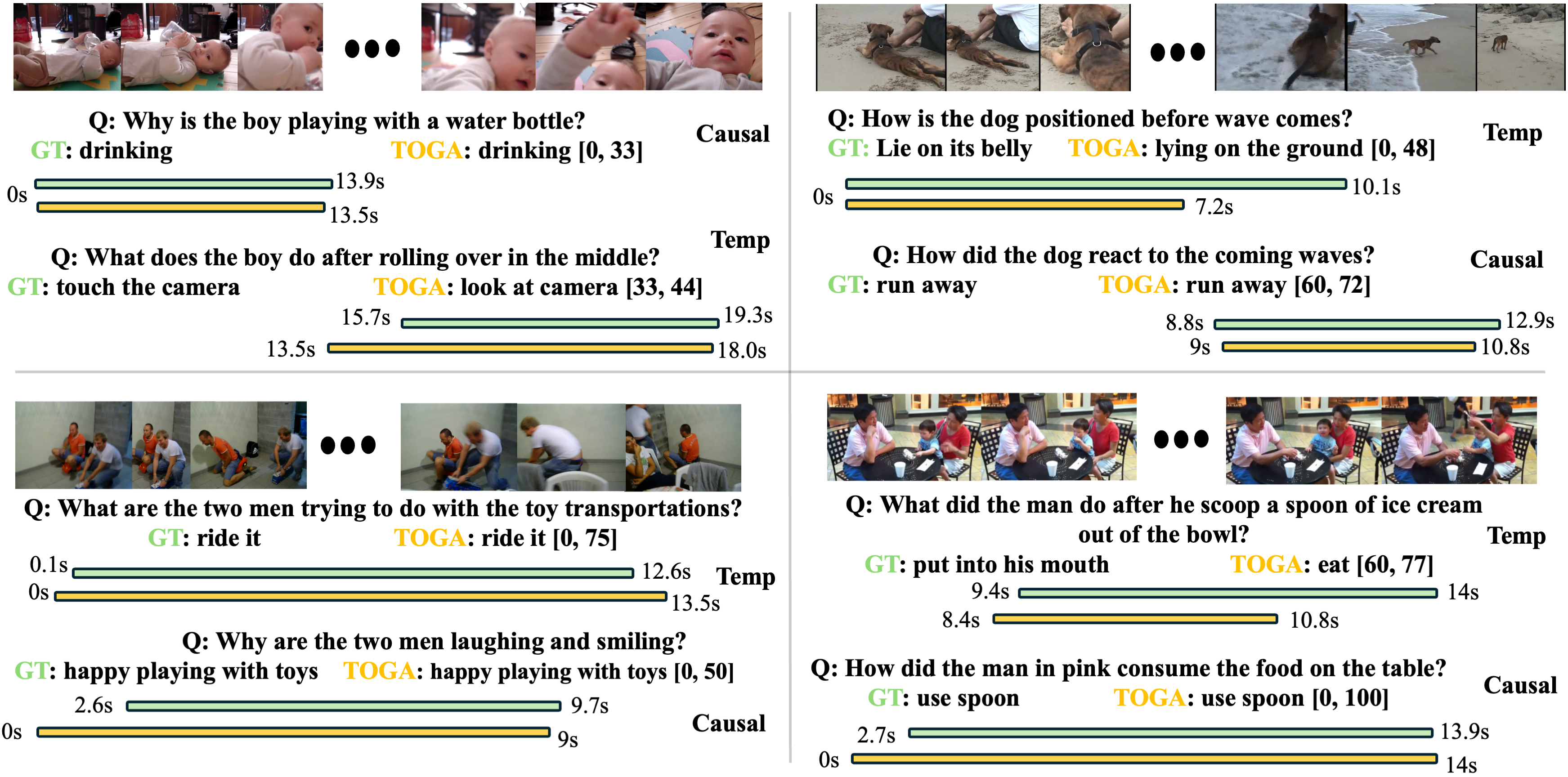}
    }
    \caption{Qualitative results on NExT-GQA. We present the ground truth and predicted answers with groundings. Ground truth segments are marked in green and predicted segments are marked in yellow. A causal and a temporal question are selected from a set of questions for each video. 
    \approach mostly generates correct answers and grounding. In some cases, predictions are different from the ground truth (GT) answers. This can be attributed to the open-ended nature of our approach.
    \vspace{-3mm}
    }
    \label{fig:qualitative-nextgqa}
\end{figure*}

\begin{figure}
    \centering
    \resizebox{0.95\linewidth}{!}{
        \includegraphics{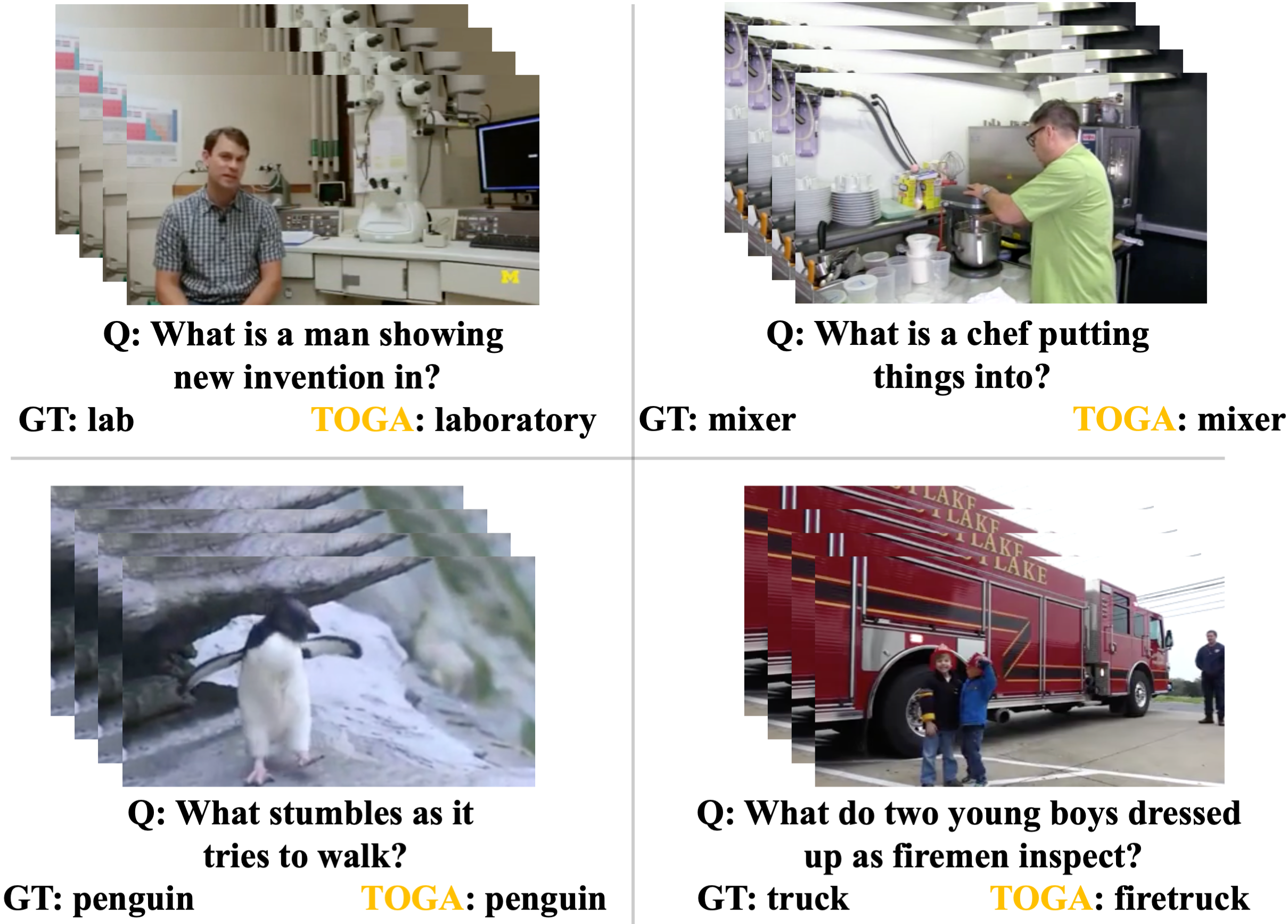}
    }
    \caption{Qualitative results for open-ended videoQA on MSVD. \approach can generate correct but slightly different answers from the ground truth due to its open-ended nature.
    \vspace{-4mm}
    }
    \label{fig:msvd-success}
\end{figure}

\textbf{Qualitative results.}
%\AG{Can mention avg len of GT vs pred here. Do detailed analysis in supp.}
We include some qualitative examples from the NExT-GQA dataset in \cref{fig:qualitative-nextgqa}. We can observe that due to the open-ended nature of our setup and because of the fact that the model is not seeing the options to the question, the model can generate responses similar in meaning to, but not matching, the ground truth exactly. 
Along with generating the answer, our method is able to simultaneously ground the answer. 
Each video in NExT-GQA has multiple questions, and we show examples of one temporal and one causal query for each video shown here. We also include  qualitative examples taken from the MSVD-QA dataset in \cref{fig:msvd-success}. Here as well, the model may generate answers that do not match the ground truth exactly but might convey the same meaning. 

%\vspace{-3mm}
\subsection{Ablation and analysis}
%\AG{Interleaving ablation}
We perform ablations on major components of our approach and analyze the results on various experimental setups.

\textbf{Ablation on multi-scale vision-language connector.}
Recall that our MS-VLC module processes videos at two temporal resolutions: 16 frames at the dense scale and 4 frames at the sparse scale. 
To justify the importance of this module, we consider frameworks with only one temporal resolution and evaluate on NExT-GQA~\cite{nextgqa-dataset}. 
To better understand the effect of this module, we divide the answers into three groups based on the length of the true grounding: 1) short events with a grounding length of less than 30\% of the video, 2) medium events with a grounding length of 30-70\% of the video, and 3) long events with a grounding length of higher than 70\% of the video. 
We present the IoU metric for these in \cref{tab:sparse-dense}. 
Note that the multi-scale model performs better than single-scale models and the effect is more prominent for answers with short and long durations.

%We observe that the sparse branch with lower temporal resolution is able to ground larger events better. However, the dense branch is able to perform better on queries relevant to smaller, fine-grained sections of the video. 

% \usepackage{graphicx}

\begin{table}
\centering
\resizebox{\linewidth}{!}{%
\begin{tabular}{c|c|ccc}
 & \multicolumn{4}{c}{Query Type} \\
Model type & All & short & medium & long \\ 
\hline
Sparse
  only & 20.0 & 16.2 & 28.9 & 47.5 \\
Dense
  Only & 22.1 & 18.3 & 32.2 & 32.1 \\
Multi-Scale (MS-VLC) & \textbf{24.4} & \textbf{20.5} & \textbf{34.7} & \textbf{49.3}
\end{tabular}
}
\caption{\label{tab:sparse-dense}
The MS-VLC effectively improves grounding performance. This effect is more prominent for answers with short and long temporal durations. 
\vspace{-2mm}
}

\end{table}

\textbf{Ablation on the consistency constraint.}
%
% \AG{No table - just mention that no timestamp predicted after pretraining. After IT stage, mIoU is around 12. After Consistency, mIoU is 24}
For weakly supervised grounding, at the final stage of training, we impose a consistency constraint between the grounding response and the response generated by a question with the same grounding. 
To justify this, we train a variant without this stage of training where we instruction tune the model only with pseudo temporal groundings. This results in an mIoU of 12.1 which is significantly low compared to the mIoU of 24.4 on NExT-GQA with the final stage of training.

% The pretraining stage for training the vision-language connector is a common approach used to align vision and language domains \cite{mm1-tips-llm, meta-intro-vlm}.
% This stage teaches the LLM to understand the visual content in the input. 
% The IT stage trains the model to perform QA and understand the concept of temporal timestamps. While our instruction-tuned model can perform grounding, the quality of these timestamps is not very good, as can be seen in \AG{Table}. 
% A more interesting observation about the model after the IT stage is that it does not generate a timestamp every time it is prompted to. When we evaluate the model post IT stage on NExT-GQA, it only generates a timestamp for \AG{x} percentage of queries on the test set. 
% However, once we perform the consistency stage training, the predicted timestamp quality improves and the model learns to predict the timestamp every time. 

% \AG{Doing IT and consistency together - we tried, but pretrained model does not generate timestamp - Do this experiment!}

\textbf{Analysis of the type of questions.}
%
%\AG{MS-VLC ablation on question types?}
NExT-GQA consists of two primary types of questions: causal and temporal queries. Causal questions include why and how questions. 
Temporal questions are divided into past, present, and future types based on the timing of the answer. 
For example, \texttt{What did the boy do after standing up} is an example of a future temporal question. 
Both of these question types involve different levels of reasoning. We present the results corresponding to the Acc@GQA metric in \cref{tab:nextgqa-question-type}. 
We notice that the temporal questions are more difficult than causal questions as they require understanding the sequence of events. 
Temporal questions, especially those referring to the past or future, e.g. \texttt{What did the boy do \textit{after} he stood up from the ground?} 
%\RC{change question here, same as in intro?}
require more long-term reasoning to generate correct answers with grounding.

\begin{table}
\centering
\resizebox{0.9\linewidth}{!}{%
\begin{tabular}{c|cc|ccc}
\multirow{2}{*}{Question types} & \multicolumn{2}{c|}{Causal} & \multicolumn{3}{c}{Temporal} \\
 & Why & How & Present & Past & Future \\ 
\hline
Acc@GQA~ & 26.1 & 27.4 & 23.4 & 18 & 18.1
\end{tabular}
}
\caption{\label{tab:nextgqa-question-type}
Acc@GQA for causal and temporal question types on NeXT-GQA. We observe that temporal questions are more difficult, specifically questions referring to past and future events. 
%of a particular event - since they require more long-term reasoning.  
}
\end{table}

\textbf{Analysis of the number of frames.}
We explore the effect of the frame count on the grounding performance on NExT-GQA. We choose a setup (16, 4), i.e., 16 frames for the dense and 4 for the sparse scale.
We experiment with other setups with lower frames (8, 2) and higher frames (32, 8).
The (8, 2) setup achieves a mIoU of 20.8. The (32, 8) setup achieves a better mIoU of 21.5 with a three times higher training time.
Both are lower than the mIoU of 24.4 with (16, 4). We believe more frames better encode the video features, but may introduce spurious features and make it harder to train the model. 
More combinations, along with a variant with frame indices in the prompt, are in the supplementary material.
%Next, we input the frame indices in the prompt. This is discussed in supplementary material.

% Moreover, grounding for these temporal queries is also more challenging due to different activities being involved in the question and answer, eg. do you ground the moment the boy stands up, or do you ground the moment he does the next activity? 
% Answering and grounding such queries is challenging and hence, the Acc@GQA metric, considering both these tasks simultaneously, is lower for temporal queries.

\textbf{Analysis of temporal representation.}
We consider a temporal representation of [$<$start$>$, $<$end$>$] where start and end are integers $\in [0, 100]$, with 0 marking the start of the video and 100 marking the end. 
We chose this range based on the length of the videos in seconds. We scale the $[0, 100]$ range to video start and end times in seconds. 
We also considered a scenario where the start and end are floats $\in [0, 1]$. We achieve a lower mIoU of 19.0 on NeXT-GQA with the [0, 1] range, compared to 24.4 with [0,100]. 
We notice it is harder to learn the floating point representation with language tokens. A similar behavior is shown in \cite{mr-blip-grounding-llm}. 

% \textbf{Representation of time}
% We choose to represent time as integers from 0 to 100, with 0 marking the beginning of the video and 100 marking the end of the video. 
% We train the model to output timestamps relative to the video in this range, quantize the predictions to integers. We also experiment with representing time as floating point numbers in the range [0,1]. We observe that this model achieves 19.0 mIoU on NeXT-GQA, as opposed to 24.4 when using the range [0,100]. We conclude that quantizing the timestamps in a larger range is a simpler representation of time for the model, potentially because floating point numbers require more tokens to be generated when compared to integers. Our observations match those made by \cite{mr-blip-grounding-llm}.

% \begin{figure*}[htbp]
% \centering
% \begin{subfigure}{0.49\textwidth}
% \includegraphics[width=\textwidth]{figures/Failure case dog on beach.png}
% \caption{\label{fig:failure-case-dog}}
% \end{subfigure}
% \hfill
% \begin{subfigure}{0.49\textwidth}
% \includegraphics[width=\textwidth]{figures/Failure case robot.png}
% \caption{\label{fig:failure-case-robot}}
% \end{subfigure}
% \caption{Some qualitative examples from the NExT-GQA dataset. Successes are shown in \textcolor{ForestGreen}{\textbf{green}} boxes, while failure cases are in \textcolor{red}{\textbf{red}} boxes.
% Due to its open-ended nature, \approach can generate answers which might be correct but unrelated to the pre-defined options, and can be penalized for it. \AG{Show options? Replace pred: with VIGOR: }
% }
% \label{fig:overall}
% \end{figure*}

\begin{figure}
    \centering
    \resizebox{0.95\linewidth}{!}{
        \includegraphics{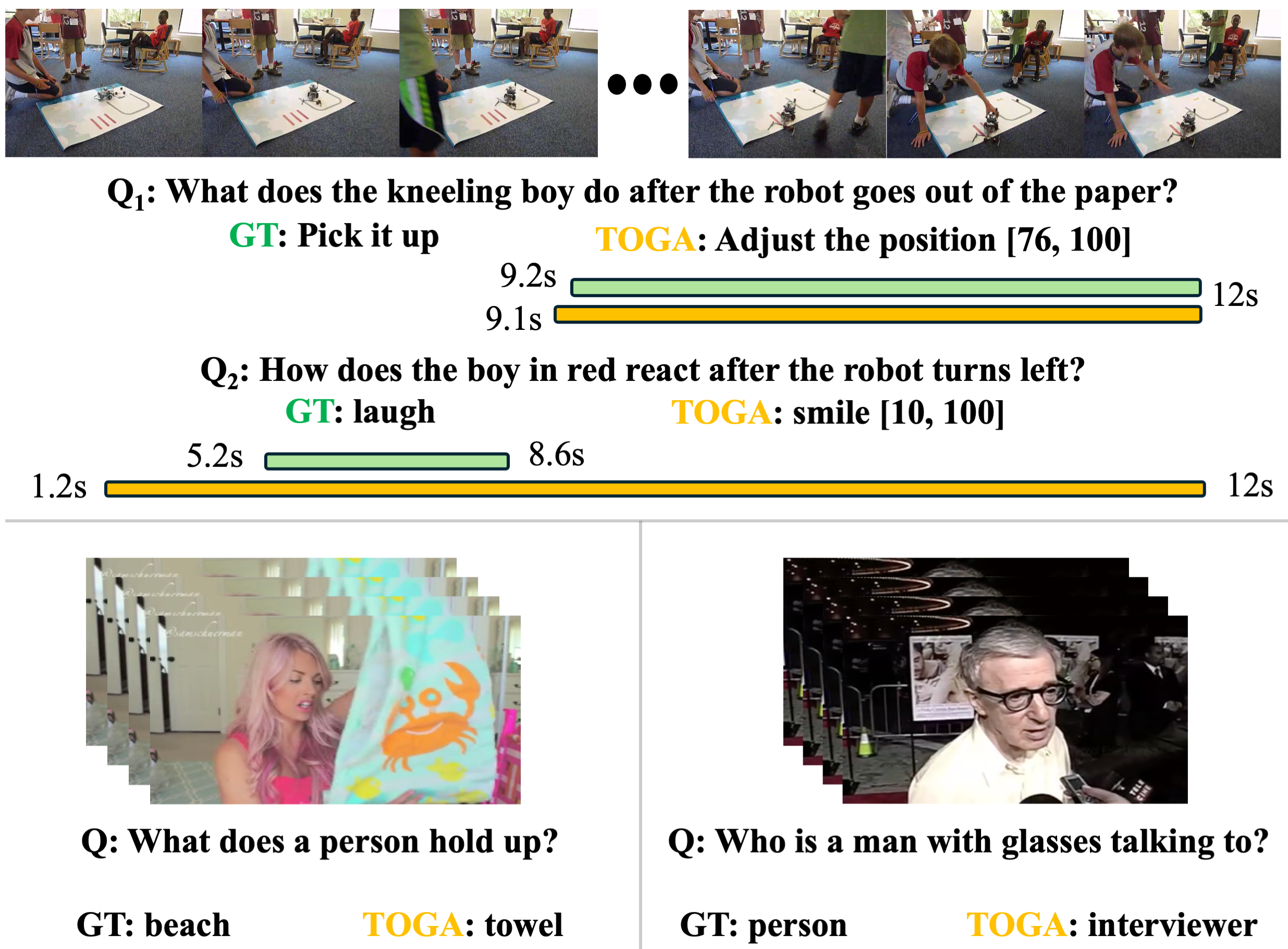}
    }
    \caption{
    The top part presents an example from NExT-GQA where \approach fails to ground the answer. The bottom part presents two examples from MSVD. 
    \approach generates answers that may be relevant but do not exactly match the GT.
    \vspace{-2mm}
    % \AG{MSVD second example, GT and VIGOR too close. Susmit: I agree .. could we find something new or just use one and save space?}
    }
    \label{fig:failure-cases}
\end{figure}

\textbf{Failure cases.} 
We present the failure cases on grounding and QA tasks in Fig.~\ref{fig:failure-cases}. On the top, we present a case from NExT-GQA where \approach's response is somewhat relevant but does not match the ground truth. This is due to the open-ended answer generation. 
In another instance, we generate a closely matched answer \texttt{smile} vs. \texttt{laugh}, but the grounding is inaccurate. This is due to the weakly supervised learning where we train with noisy grounding labels. 
At the bottom, we present two examples from MSVD. \approach generates a suitable answer of \texttt{towel} in response to \texttt{what does a person hold up} which is different from the ground truth. In the other case, the response \texttt{interviewer} is a special type of \texttt{person}.

\textbf{Limitations.}
We generate pseudo labels to train \approach to perform grounding in a weakly supervised setup. We consider one grounding interval for each answer. Thus, we cannot handle cases where the evidence for an answer is distributed across multiple intervals. 
We also answer all questions for a video independently. 
However, capturing the temporal dependencies between the answers could help generate more accurate answers with groundings.
This is particularly helpful for the temporal questions with before/after clauses on NExT-GQA. These are the future directions for \approach.

\vspace{-2mm}
\section{Conclusion}

We have presented \approach, a vision-language model for open-ended video QA with temporal grounding. We operate in a weakly supervised setup and do not rely on temporal annotations. \approach consists of an MS-VLC module to capture both high-frequency and low-frequency temporal features. We have instruction tuned the MS-VLC and language decoder to jointly generate open-ended answers with temporal grounding. 
Unlike existing approaches, we do not require the options to generate an answer. Our experiments show that jointly generating grounded answers improves the accuracy of both answers and temporal grounding. 
We have evaluated \approach on NExT-GQA for grounded QA and MSVD-QA and ActivityNet-QA for open-ended QA. We achieve state-of-the-art performance on these benchmarks. 

% Our ablation study shows that the MS-VLC module is effective at grounding a wide length of temporal segments.

\section*{Acknowledgment}
This work was supported in part by the U.S. Air Force and DARPA under Contract No. FA8750-23-C-0519, the U.S. Army Research Laboratory Cooperative Research Agreement W911NF-17-2-0196, the U.S. Army Combat Capabilities Development Command (DEVCOM)
Army Research Laboratory under Support Agreement No. USMA 21050, and the DARPA under Support Agreement No. USMA 23004. The opinions, findings, and conclusions expressed in this paper are those of the authors and do not reflect the position of the United States Department of Defense or the United States Government.

%\clearpage
{
    \small
    \bibliographystyle{ieeenat_fullname}
    \bibliography{iccv_main.bib}

\begin{thebibliography}{60}
\providecommand{\natexlab}[1]{#1}
\providecommand{\url}[1]{\texttt{#1}}
\expandafter\ifx\csname urlstyle\endcsname\relax
  \providecommand{\doi}[1]{doi: #1}\else
  \providecommand{\doi}{doi: \begingroup \urlstyle{rm}\Url}\fi

\bibitem[Antol et~al.(2015)Antol, Agrawal, Lu, Mitchell, Batra, Zitnick, and Parikh]{vqa-1}
Stanislaw Antol, Aishwarya Agrawal, Jiasen Lu, Margaret Mitchell, Dhruv Batra, C.~Lawrence Zitnick, and Devi Parikh.
\newblock Vqa: Visual question answering.
\newblock In \emph{Proceedings of the IEEE International Conference on Computer Vision (ICCV)}, 2015.

\bibitem[Chen and Dolan(2011)]{msvd}
David Chen and William Dolan.
\newblock Collecting highly parallel data for paraphrase evaluation.
\newblock In \emph{Proceedings of the 49th Annual Meeting of the Association for Computational Linguistics: Human Language Technologies}, pages 190--200, Portland, Oregon, USA, 2011. Association for Computational Linguistics.

\bibitem[Chen et~al.(2024{\natexlab{a}})Chen, Liao, Lin, Yu, Chen, and Wang]{rextime}
Jr-Jen Chen, Yu-Chien Liao, Hsi-Che Lin, Yu-Chu Yu, Yen-Chun Chen, and Yu-Chiang~Frank Wang.
\newblock Rextime: A benchmark suite for reasoning-across-time in videos.
\newblock \emph{arXiv preprint arXiv:2406.19392}, 2024{\natexlab{a}}.

\bibitem[Chen et~al.(2024{\natexlab{b}})Chen, Siarohin, Menapace, Deyneka, wei Chao, Jeon, Fang, Lee, Ren, Yang, and Tulyakov]{panda70m}
Tsai-Shien Chen, Aliaksandr Siarohin, Willi Menapace, Ekaterina Deyneka, Hsiang wei Chao, Byung~Eun Jeon, Yuwei Fang, Hsin-Ying Lee, Jian Ren, Ming-Hsuan Yang, and Sergey Tulyakov.
\newblock Panda-70m: Captioning 70m videos with multiple cross-modality teachers, 2024{\natexlab{b}}.

\bibitem[Cheng et~al.(2024)Cheng, Leng, Zhang, Xin, Li, Chen, Zhu, Zhang, Luo, Zhao, and Bing]{videollama2}
Zesen Cheng, Sicong Leng, Hang Zhang, Yifei Xin, Xin Li, Guanzheng Chen, Yongxin Zhu, Wenqi Zhang, Ziyang Luo, Deli Zhao, and Lidong Bing.
\newblock Videollama 2: Advancing spatial-temporal modeling and audio understanding in video-llms.
\newblock \emph{arXiv preprint arXiv:2406.07476}, 2024.

\bibitem[Fabian Caba~Heilbron and Niebles(2015)]{activitynet}
Bernard~Ghanem Fabian Caba~Heilbron, Victor~Escorcia and Juan~Carlos Niebles.
\newblock Activitynet: A large-scale video benchmark for human activity understanding.
\newblock In \emph{Proceedings of the IEEE Conference on Computer Vision and Pattern Recognition}, pages 961--970, 2015.

\bibitem[Fang et~al.(2020)Fang, Gokhale, Banerjee, Baral, and Yang]{kbvqa-2}
Zhiyuan Fang, Tejas Gokhale, Pratyay Banerjee, Chitta Baral, and Yezhou Yang.
\newblock Video2commonsense: Generating commonsense descriptions to enrich video captioning.
\newblock \emph{arXiv preprint arXiv:2003.05162}, 2020.

\bibitem[Feichtenhofer et~al.(2019)Feichtenhofer, Fan, Malik, and He]{feichtenhofer2019slowfast}
Christoph Feichtenhofer, Haoqi Fan, Jitendra Malik, and Kaiming He.
\newblock Slowfast networks for video recognition.
\newblock In \emph{Proceedings of the IEEE/CVF international conference on computer vision}, pages 6202--6211, 2019.

\bibitem[Garcia et~al.(2020)Garcia, Otani, Chu, and Nakashima]{kbvqa-1}
Noa Garcia, Mayu Otani, Chenhui Chu, and Yuta Nakashima.
\newblock Knowit vqa: Answering knowledge-based questions about videos.
\newblock In \emph{Proceedings of the AAAI conference on artificial intelligence}, pages 10826--10834, 2020.

\bibitem[Goyal et~al.(2019)Goyal, Khot, Agrawal, Summers-Stay, Batra, and Parikh]{vqa-2}
Yash Goyal, Tejas Khot, Aishwarya Agrawal, Douglas Summers-Stay, Dhruv Batra, and Devi Parikh.
\newblock Making the v in vqa matter: Elevating the role of image understanding in visual question answering.
\newblock \emph{Int. J. Comput. Vision}, 127\penalty0 (4):\penalty0 398–414, 2019.

\bibitem[Heilman and Smith(2009)]{heilman2009question}
Michael Heilman and Noah~A Smith.
\newblock Question generation via overgenerating transformations and ranking.
\newblock \emph{DTIC Document}, 2009.

\bibitem[Huang et~al.(2024)Huang, Liao, Radhakrishnan, Yin, Molchanov, Yu, and Kautz]{lita}
De-An Huang, Shijia Liao, Subhashree Radhakrishnan, Hongxu Yin, Pavlo Molchanov, Zhiding Yu, and Jan Kautz.
\newblock Lita: Language instructed temporal-localization assistant.
\newblock In \emph{Computer Vision – ECCV 2024: 18th European Conference, Milan, Italy, September 29–October 4, 2024, Proceedings, Part LXIV}, page 202–218, Berlin, Heidelberg, 2024. Springer-Verlag.

\bibitem[Jang et~al.(2017)Jang, Song, Yu, Kim, and Kim]{jang2017tgif}
Yunseok Jang, Yale Song, Youngjae Yu, Youngjin Kim, and Gunhee Kim.
\newblock Tgif-qa: Toward spatio-temporal reasoning in visual question answering.
\newblock In \emph{Proceedings of the IEEE conference on computer vision and pattern recognition}, pages 2758--2766, 2017.

\bibitem[Jang et~al.(2019{\natexlab{a}})Jang, Song, Kim, Yu, Kim, and Kim]{tgif-qa}
Yunseok Jang, Yale Song, Chris~Dongjoo Kim, Youngjae Yu, Youngjin Kim, and Gunhee Kim.
\newblock Video question answering with spatio-temporal reasoning.
\newblock \emph{IJCV}, 2019{\natexlab{a}}.

\bibitem[Jang et~al.(2019{\natexlab{b}})Jang, Song, Kim, Yu, Kim, and Kim]{video-vqa-2}
Yunseok Jang, Yale Song, Chris~Dongjoo Kim, Youngjae Yu, Youngjin Kim, and Gunhee Kim.
\newblock Video question answering with spatio-temporal reasoning.
\newblock \emph{Int. J. Comput. Vision}, 127\penalty0 (10):\penalty0 1385–1412, 2019{\natexlab{b}}.

\bibitem[Jiang et~al.(2023)Jiang, Sablayrolles, Mensch, Bamford, Chaplot, de~las Casas, Bressand, Lengyel, Lample, Saulnier, Lavaud, Lachaux, Stock, Scao, Lavril, Wang, Lacroix, and Sayed]{mistral-7b}
Albert~Q. Jiang, Alexandre Sablayrolles, Arthur Mensch, Chris Bamford, Devendra~Singh Chaplot, Diego de~las Casas, Florian Bressand, Gianna Lengyel, Guillaume Lample, Lucile Saulnier, Lélio~Renard Lavaud, Marie-Anne Lachaux, Pierre Stock, Teven~Le Scao, Thibaut Lavril, Thomas Wang, Timothée Lacroix, and William~El Sayed.
\newblock Mistral 7b, 2023.

\bibitem[Jin et~al.(2023)Jin, Takanobu, Zhang, Cao, and Yuan]{chat-univi}
Peng Jin, Ryuichi Takanobu, Caiwan Zhang, Xiaochun Cao, and Li Yuan.
\newblock Chat-univi: Unified visual representation empowers large language models with image and video understanding.
\newblock \emph{arXiv preprint arXiv:2311.08046}, 2023.

\bibitem[Kazakos et~al.(2021)Kazakos, Nagrani, Zisserman, and Damen]{kazakos2021slow}
Evangelos Kazakos, Arsha Nagrani, Andrew Zisserman, and Dima Damen.
\newblock Slow-fast auditory streams for audio recognition.
\newblock In \emph{ICASSP 2021-2021 IEEE International Conference on Acoustics, Speech and Signal Processing (ICASSP)}, pages 855--859. IEEE, 2021.

\bibitem[Ko et~al.(2023)Ko, Lee, Choi, Chu, Park, and Kim]{ovqa-benchmark}
Dohwan Ko, Ji~Soo Lee, Miso Choi, Jaewon Chu, Jihwan Park, and Hyunwoo~J Kim.
\newblock Open-vocabulary video question answering: A new benchmark for evaluating the generalizability of video question answering models.
\newblock In \emph{Proceedings of the IEEE/CVF international conference on computer vision}, 2023.

\bibitem[Krishna et~al.(2017)Krishna, Hata, Ren, Fei-Fei, and Niebles]{activitynet-captions}
Ranjay Krishna, Kenji Hata, Frederic Ren, Li Fei-Fei, and Juan~Carlos Niebles.
\newblock Dense-captioning events in videos, 2017.

\bibitem[Lavie and Agarwal(2007)]{meteor}
Alon Lavie and Abhaya Agarwal.
\newblock Meteor: an automatic metric for mt evaluation with high levels of correlation with human judgments.
\newblock In \emph{Proceedings of the Second Workshop on Statistical Machine Translation}, page 228–231, USA, 2007. Association for Computational Linguistics.

\bibitem[Lei et~al.(2019)Lei, Yu, Bansal, and Berg]{tvqa}
Jie Lei, Licheng Yu, Mohit Bansal, and Tamara~L. Berg.
\newblock Tvqa: Localized, compositional video question answering, 2019.

\bibitem[Lei et~al.(2020)Lei, Yu, Berg, and Bansal]{tvqa-plus}
Jie Lei, Licheng Yu, Tamara~L. Berg, and Mohit Bansal.
\newblock Tvqa+: Spatio-temporal grounding for video question answering, 2020.

\bibitem[Li et~al.(2023)Li, He, Wang, Li, Wang, Luo, Wang, Wang, and Qiao]{videochat}
Kunchang Li, Yinan He, Yi Wang, Yizhuo Li, Wenhai Wang, Ping Luo, Yali Wang, Limin Wang, and Yu Qiao.
\newblock Videochat: Chat-centric video understanding.
\newblock \emph{arXiv preprint arXiv:2305.06355}, 2023.

\bibitem[Li et~al.(2022)Li, Wang, Xiao, and Chua]{igv}
Yicong Li, Xiang Wang, Junbin Xiao, and Tat-Seng Chua.
\newblock Equivariant and invariant grounding for video question answering.
\newblock In \emph{Proceedings of the 30th ACM International Conference on Multimedia}, page 4714–4722, New York, NY, USA, 2022. Association for Computing Machinery.

\bibitem[Lin et~al.(2023)Lin, Zhu, Ye, Ning, Jin, and Yuan]{videollava}
Bin Lin, Bin Zhu, Yang Ye, Munan Ning, Peng Jin, and Li Yuan.
\newblock Video-llava: Learning united visual representation by alignment before projection.
\newblock \emph{arXiv preprint arXiv:2311.10122}, 2023.

\bibitem[Liu et~al.(2023)Liu, Li, Wu, and Lee]{llava}
Haotian Liu, Chunyuan Li, Qingyang Wu, and Yong~Jae Lee.
\newblock Visual instruction tuning, 2023.

\bibitem[Loshchilov(2017)]{loshchilov2017decoupled}
I Loshchilov.
\newblock Decoupled weight decay regularization.
\newblock \emph{arXiv preprint arXiv:1711.05101}, 2017.

\bibitem[Maaz et~al.(2024)Maaz, Rasheed, Khan, and Khan]{videochatgpt}
Muhammad Maaz, Hanoona Rasheed, Salman Khan, and Fahad~Shahbaz Khan.
\newblock Video-chatgpt: Towards detailed video understanding via large vision and language models, 2024.

\bibitem[Meinardus et~al.(2024)Meinardus, Batra, Rohrbach, and Rohrbach]{mr-blip-grounding-llm}
Boris Meinardus, Anil Batra, Anna Rohrbach, and Marcus Rohrbach.
\newblock The surprising effectiveness of multimodal large language models for video moment retrieval, 2024.

\bibitem[Micheletti et~al.(2024)Micheletti, Belkadi, Han, and Nenadic]{causal-lm}
Nicolo Micheletti, Samuel Belkadi, Lifeng Han, and Goran Nenadic.
\newblock Exploration of masked and causal language modelling for text generation, 2024.

\bibitem[Miller(1995)]{wordnet}
George~A. Miller.
\newblock Wordnet: a lexical database for english.
\newblock \emph{Commun. ACM}, 38\penalty0 (11):\penalty0 39–41, 1995.

\bibitem[OpenAI(2024)]{gpt4}
OpenAI.
\newblock Gpt-4 technical report, 2024.

\bibitem[Qian et~al.(2025)Qian, Dong, Zhang, Zang, Ding, Lin, and Wang]{videostreaming}
Rui Qian, Xiaoyi Dong, Pan Zhang, Yuhang Zang, Shuangrui Ding, Dahua Lin, and Jiaqi Wang.
\newblock Streaming long video understanding with large language models.
\newblock \emph{Advances in Neural Information Processing Systems}, 37:\penalty0 119336--119360, 2025.

\bibitem[Radford et~al.(2021)Radford, Kim, Hallacy, Ramesh, Goh, Agarwal, Sastry, Askell, Mishkin, Clark, Krueger, and Sutskever]{CLIP}
Alec Radford, Jong~Wook Kim, Chris Hallacy, Aditya Ramesh, Gabriel Goh, Sandhini Agarwal, Girish Sastry, Amanda Askell, Pamela Mishkin, Jack Clark, Gretchen Krueger, and Ilya Sutskever.
\newblock Learning transferable visual models from natural language supervision, 2021.

\bibitem[Radosavovic et~al.(2020)Radosavovic, Kosaraju, Girshick, He, and Dollar]{regnet}
Ilija Radosavovic, Raj~Prateek Kosaraju, Ross Girshick, Kaiming He, and Piotr Dollar.
\newblock Designing network design spaces.
\newblock In \emph{Proceedings of the IEEE/CVF Conference on Computer Vision and Pattern Recognition (CVPR)}, 2020.

\bibitem[Speer et~al.(2017)Speer, Chin, and Havasi]{conceptnet}
Robyn Speer, Joshua Chin, and Catherine Havasi.
\newblock Conceptnet 5.5: an open multilingual graph of general knowledge.
\newblock In \emph{Proceedings of the Thirty-First AAAI Conference on Artificial Intelligence}, page 4444–4451. AAAI Press, 2017.

\bibitem[Sur{\'\i}s et~al.(2023)Sur{\'\i}s, Menon, and Vondrick]{vipergpt}
D{\'\i}dac Sur{\'\i}s, Sachit Menon, and Carl Vondrick.
\newblock Vipergpt: Visual inference via python execution for reasoning.
\newblock In \emph{Proceedings of the IEEE/CVF International Conference on Computer Vision}, pages 11888--11898, 2023.

\bibitem[Tapaswi et~al.(2016)Tapaswi, Zhu, Stiefelhagen, Torralba, Urtasun, and Fidler]{mmvqa-1}
Makarand Tapaswi, Yukun Zhu, Rainer Stiefelhagen, Antonio Torralba, Raquel Urtasun, and Sanja Fidler.
\newblock Movieqa: Understanding stories in movies through question-answering.
\newblock In \emph{Proceedings of the IEEE conference on computer vision and pattern recognition}, pages 4631--4640, 2016.

\bibitem[Team(2024)]{gemini}
Gemini Team.
\newblock Gemini: A family of highly capable multimodal models, 2024.

\bibitem[Touvron et~al.(2023)Touvron, Lavril, Izacard, Martinet, Lachaux, Lacroix, Rozière, Goyal, Hambro, Azhar, Rodriguez, Joulin, Grave, and Lample]{llama}
Hugo Touvron, Thibaut Lavril, Gautier Izacard, Xavier Martinet, Marie-Anne Lachaux, Timothée Lacroix, Baptiste Rozière, Naman Goyal, Eric Hambro, Faisal Azhar, Aurelien Rodriguez, Armand Joulin, Edouard Grave, and Guillaume Lample.
\newblock Llama: Open and efficient foundation language models, 2023.

\bibitem[Wang et~al.(2024{\natexlab{a}})Wang, Xu, Cheng, Diao, Zhou, Cao, Wang, Ge, and Huang]{groundedvideollm}
Haibo Wang, Zhiyang Xu, Yu Cheng, Shizhe Diao, Yufan Zhou, Yixin Cao, Qifan Wang, Weifeng Ge, and Lifu Huang.
\newblock Grounded-videollm: Sharpening fine-grained temporal grounding in video large language models, 2024{\natexlab{a}}.

\bibitem[Wang et~al.(2024{\natexlab{b}})Wang, Zhang, Zohar, and Yeung-Levy]{videoagent}
Xiaohan Wang, Yuhui Zhang, Orr Zohar, and Serena Yeung-Levy.
\newblock Videoagent: Long-form video understanding with large language model as agent.
\newblock \emph{arXiv preprint arXiv:2403.10517}, 2024{\natexlab{b}}.

\bibitem[Wang et~al.(2024{\natexlab{c}})Wang, Yu, Stengel-Eskin, Yoon, Cheng, Bertasius, and Bansal]{videotree}
Ziyang Wang, Shoubin Yu, Elias Stengel-Eskin, Jaehong Yoon, Feng Cheng, Gedas Bertasius, and Mohit Bansal.
\newblock Videotree: Adaptive tree-based video representation for llm reasoning on long videos.
\newblock \emph{arXiv preprint arXiv:2405.19209}, 2024{\natexlab{c}}.

\bibitem[Wu et~al.(2024{\natexlab{a}})Wu, Yu, Chen, Tenenbaum, and Gan]{wu2024star}
Bo Wu, Shoubin Yu, Zhenfang Chen, Joshua~B Tenenbaum, and Chuang Gan.
\newblock Star: A benchmark for situated reasoning in real-world videos.
\newblock \emph{arXiv preprint arXiv:2405.09711}, 2024{\natexlab{a}}.

\bibitem[Wu et~al.(2024{\natexlab{b}})Wu, Hu, Sun, Zhou, Zhu, Rao, Schiele, and Yang]{numberit}
Yongliang Wu, Xinting Hu, Yuyang Sun, Yizhou Zhou, Wenbo Zhu, Fengyun Rao, Bernt Schiele, and Xu Yang.
\newblock Number it: Temporal grounding videos like flipping manga, 2024{\natexlab{b}}.

\bibitem[Xiao et~al.(2020)Xiao, Lee, Grauman, Malik, and Feichtenhofer]{xiao2020audiovisual}
Fanyi Xiao, Yong~Jae Lee, Kristen Grauman, Jitendra Malik, and Christoph Feichtenhofer.
\newblock Audiovisual slowfast networks for video recognition.
\newblock \emph{arXiv preprint arXiv:2001.08740}, 2020.

\bibitem[Xiao et~al.(2021)Xiao, Shang, Yao, and Chua]{nextqa}
Junbin Xiao, Xindi Shang, Angela Yao, and Tat-Seng Chua.
\newblock Next-qa:next phase of question-answering to explaining temporal actions, 2021.

\bibitem[Xiao et~al.(2024{\natexlab{a}})Xiao, Huang, Qin, Li, Li, Zhu, Tao, Yu, Lin, Chua, and Yao]{videoqa-llm-survey}
Junbin Xiao, Nanxin Huang, Hangyu Qin, Dongyang Li, Yicong Li, Fengbin Zhu, Zhulin Tao, Jianxing Yu, Liang Lin, Tat-Seng Chua, and Angela Yao.
\newblock Videoqa in the era of llms: An empirical study, 2024{\natexlab{a}}.

\bibitem[Xiao et~al.(2024{\natexlab{b}})Xiao, Yao, Li, and Chua]{nextgqa-dataset}
Junbin Xiao, Angela Yao, Yicong Li, and Tat-Seng Chua.
\newblock Can i trust your answer? visually grounded video question answering.
\newblock In \emph{Proceedings of the IEEE/CVF Conference on Computer Vision and Pattern Recognition}, pages 13204--13214, 2024{\natexlab{b}}.

\bibitem[Xu et~al.(2017{\natexlab{a}})Xu, Zhao, Xiao, Wu, Zhang, He, and Zhuang]{msvd-msrvttqa}
Dejing Xu, Zhou Zhao, Jun Xiao, Fei Wu, Hanwang Zhang, Xiangnan He, and Yueting Zhuang.
\newblock Video question answering via gradually refined attention over appearance and motion.
\newblock In \emph{ACM Multimedia}, 2017{\natexlab{a}}.

\bibitem[Xu et~al.(2017{\natexlab{b}})Xu, Zhao, Xiao, Wu, Zhang, He, and Zhuang]{xu2017video}
Dejing Xu, Zhou Zhao, Jun Xiao, Fei Wu, Hanwang Zhang, Xiangnan He, and Yueting Zhuang.
\newblock Video question answering via gradually refined attention over appearance and motion.
\newblock In \emph{Proceedings of the 25th ACM international conference on Multimedia}, pages 1645--1653, 2017{\natexlab{b}}.

\bibitem[Yang et~al.(2022{\natexlab{a}})Yang, Miech, Sivic, Laptev, and Schmid]{frozenbilm}
Antoine Yang, Antoine Miech, Josef Sivic, Ivan Laptev, and Cordelia Schmid.
\newblock Zero-shot video question answering via frozen bidirectional language models.
\newblock In \emph{NeurIPS}, 2022{\natexlab{a}}.

\bibitem[Yang et~al.(2022{\natexlab{b}})Yang, Miech, Sivic, Laptev, and Schmid]{vidstg}
Antoine Yang, Antoine Miech, Josef Sivic, Ivan Laptev, and Cordelia Schmid.
\newblock Tubedetr: Spatio-temporal video grounding with transformers.
\newblock In \emph{CVPR}, 2022{\natexlab{b}}.

\bibitem[Yu et~al.(2023)Yu, Cho, Yadav, and Bansal]{sevila}
Shoubin Yu, Jaemin Cho, Prateek Yadav, and Mohit Bansal.
\newblock Self-chained image-language model for video localization and question answering.
\newblock In \emph{NeurIPS}, 2023.

\bibitem[Yu et~al.(2019)Yu, Xu, Yu, Yu, Zhao, Zhuang, and Tao]{activitynet-qa}
Zhou Yu, Dejing Xu, Jun Yu, Ting Yu, Zhou Zhao, Yueting Zhuang, and Dacheng Tao.
\newblock Activitynet-qa: A dataset for understanding complex web videos via question answering.
\newblock In \emph{AAAI}, pages 9127--9134, 2019.

\bibitem[Zhang et~al.(2023{\natexlab{a}})Zhang, Lu, Islam, Wang, Yu, Bansal, and Bertasius]{llovi}
Ce Zhang, Taixi Lu, Md~Mohaiminul Islam, Ziyang Wang, Shoubin Yu, Mohit Bansal, and Gedas Bertasius.
\newblock A simple llm framework for long-range video question-answering, 2023{\natexlab{a}}.

\bibitem[Zhang et~al.(2023{\natexlab{b}})Zhang, Li, and Bing]{videollama}
Hang Zhang, Xin Li, and Lidong Bing.
\newblock Video-llama: An instruction-tuned audio-visual language model for video understanding.
\newblock \emph{arXiv preprint arXiv:2306.02858}, 2023{\natexlab{b}}.

\bibitem[Zhang et~al.(2023{\natexlab{c}})Zhang, Han, Liu, Gao, Zhou, Hu, Yan, Lu, Li, and Qiao]{llama-adapter}
Renrui Zhang, Jiaming Han, Chris Liu, Peng Gao, Aojun Zhou, Xiangfei Hu, Shilin Yan, Pan Lu, Hongsheng Li, and Yu Qiao.
\newblock Llama-adapter: Efficient fine-tuning of language models with zero-init attention.
\newblock \emph{arXiv preprint arXiv:2303.16199}, 2023{\natexlab{c}}.

\bibitem[Zhong et~al.(2022)Zhong, Xiao, Ji, Li, Deng, and Chua]{videoqa-survey}
Yaoyao Zhong, Junbin Xiao, Wei Ji, Yicong Li, Weihong Deng, and Tat-Seng Chua.
\newblock Video question answering: Datasets, algorithms and challenges, 2022.

\end{thebibliography}
}

\clearpage
\setcounter{page}{1}
\setcounter{section}{0}
\setcounter{table}{0}
\setcounter{figure}{0}
\maketitlesupplementary

% \section{Table 1}
% \AG{What to elaborate in table 1?}

% \section{Additional details about the approach}

% \begin{itemize}
    
% \item prompts \AG{What extra details needed?}

% %\item MS-VLC module with architecture details. (done)

% %\item implementation details with Training setting. (done)

% %\item qualitative result on nextgqa with 3-4 QA per video.

% %\item qualitative results on Activitynet

% \end{itemize}

%\section{Experiments for supple}

% \begin{itemize}

% \item MS-VLC ablation (dense/sparse) on different question types? [done]

% \item MS-VLC ablation on MSVD/ActivityNet 
%\AG{TODO, need data from bucket} 
%(Done)

% \item Single type of prompt () [Done, did not reduce performance]

%\item Different ratio of dense vs sparse sampling rates? (16:8) (Done)

%\item Failure case analysis (grounding start/end time GT vs pred) (Done)

%\item Open ended llama based results (Done)

%\item Open ended Acc@GQA metric?? using both GPT and llama? \AG{Done using gpt} 

% \item shuffle video frames during eval?

% \item Separately generate answers and grounding as ablation?

%\item Dont include output format in the chat template? How important is it?

% \end{itemize}

In the supplemental material, we provide additional implementation details, experimental results with ablation studies, and qualitative results for video question answering and grounding. 

\section{Implementation Details}

\subsection{Training Details}
The additional settings we use, including hyperparameters and implementation details, are shown in \cref{tab:impl-details-supp}. 
We train on 8x A100 GPUs, and take 7 hours each for the instruction tuning stages and 54 hours for the pretraining stage.

\subsection{Multi-scale vision language connector}
Our MS-VLC module consists of two RegNet \cite{regnet} stages along with a 3D convolution, as done in \cite{videollama2}.
RegNet, or regular networks, are convolutional network architectures drawn from the regular design space identified by \cite{regnet}. 
We take the architecture from \cite{regnet}, randomly initialize the weights, and train the RegNet and the 3D convolution from scratch over the training stages.

\section{Additional experimental analysis}

\subsection{Multi-scale vision-language connector}
In this section, we conduct ablations on the MS-VLC and analyze the performance under various conditions. Specifically, we evaluate the spare-only and dense-only models on different setups and report the effect of the multi-scale connector on performance in these tasks. 

\paragraph{Different question types}
Recall that the NExT-GQA \cite{nextgqa-dataset} dataset consists of different question types, which are broadly classified into causal and temporal. 
In this section, we evaluate the performance of the sparse-only and dense-only connectors on the different question types. We observe that MS-VLC performs the best across all  types of questions in the dataset as shown in \cref{{tab:msvlc-ques-type}}.

% \paragraph{Open-ended QA}
% In this section, we evaluate the sparse and dense only connectors on the open ended QA task, on ActivityNet-QA \cite{activitynet-qa}. The results are shown in \cref{tab:msvlc-open-qa}.

% \begin{table}
% \centering
% %\resizebox{0.\linewidth}{!}{%
% \begin{tabular}{c|c}
% Method & Open-ended accuracy \\ 
% \hline
% Sparse only & 51.1 \\
% Dense only & 51.5 \\
% MS-VLC & 52.0
% \end{tabular}
% %}
% \caption{\label{tab:msvlc-open-qa}
% Ablation on the MS-VLC for the open-ended QA task on the ActivityNet dataset. The results are obtained using GPT-assisted evaluation.
% \AG{Results almost similar?}
% }
% \end{table}

\begin{table}
\centering
\resizebox{\linewidth}{!}{%
\begin{tabular}{c|c|c|c}
\multirow{2}{*}{Config} & \multicolumn{3}{c}{Stage} \\
 & Pretraining & Instruction Tuning & Consistemcy \\ 
\hline
Vision encoder & \multicolumn{3}{c}{clip-vit-large-patch14-336} \\
Vision select layer & \multicolumn{3}{c}{-2} \\
Language Decoder & \multicolumn{3}{c}{Mistral-7B-Instruct-v0.2} \\
Optimizer & \multicolumn{3}{c}{AdamW} \\
Weight Decay & \multicolumn{3}{c}{0} \\
Deepspeed & \multicolumn{3}{c}{Zero3} \\
Epochs & \multicolumn{3}{c}{1} \\
Warmup Ratio & \multicolumn{3}{c}{0.03} \\
LR scheduler & \multicolumn{3}{c}{cosine} \\
Decoder max length & \multicolumn{3}{c}{2048} \\
Starting LR & 1e-3 & \multicolumn{2}{c}{2e-5} \\
Batch size & 256 & \multicolumn{2}{c}{128}
\end{tabular}
}
\caption{\label{tab:impl-details-supp}
Additional hyperparameter settings and other implementation details are used in our framework. 
}
\end{table}

\begin{table}
\centering
\resizebox{\linewidth}{!}{%
\begin{tabular}{c|cc|ccc|l}
\multirow{3}{*}{Connector type} & \multicolumn{5}{c|}{Question type} & \multirow{3}{*}{All} \\
 & \multicolumn{2}{c|}{Causal} & \multicolumn{3}{c|}{Temporal} &  \\ 
\cline{2-6}
 & Why & How & Present & Past & Future &  \\ 
\hline
Sparse only & 21.7 & 21.1 & 18 & 17.9 & 12.8 & \multicolumn{1}{c}{19.55} \\
Dense only & 21.8 & 22.2 & 18.8 & 18 & 13.6 & \multicolumn{1}{c}{20.12} \\
MS-VLC & 26.1 & 27.4 & 23.4 & 18 & 18.1 & \multicolumn{1}{c}{24.6}
\end{tabular}
}
\caption{\label{tab:msvlc-ques-type}
Ablation on the MS-VLC on different question types in the NExT-GQA dataset. 
We report Acc@GQA for the specific question type.
Recall that Acc@GQA considers the accuracy of both the answer and the temporal grounding.
`All' considers all the questions for calculating the metric.
}
\end{table}

\begin{table}
\centering
%\resizebox{\linewidth}{!}{%
\begin{tabular}{c|cc}
 & GT & Predictions \\ 
\hline
Mean centre position & 49.9 & 50.1 \\
Average length & 21.6 & 21.5 \\
\% of timestamps starting at 0 & 14.40\% & 21.20\% \\
\% of timestamps ending at 100 & 10.10\% & 14.05\%
\end{tabular}
%}
\caption{\label{tab:stats-gt-pred-timestamps}
Statistics of the predicted and the ground truth (GT) timestamps. All timestamps are normalized in the range [0, 100] for this analysis. 
We notice that while the mean center position and average length are similar for GT and predictions, the predictions are more biased to start at 0 and end at 100 than the GT. 
}
\end{table}

\begin{table*}
\centering
\resizebox{0.7\linewidth}{!}{%
\begin{tabular}{c|c|ccc}
\multicolumn{2}{c|}{Error type} & MS-VLC & Dense Only & Sparse only \\ 
\hline
\multirow{4}{*}{Prediction mismatch \%} & Early start \% & 24.90\% & 29.48\% & 35.28\% \\
 & Late start \%~ & 26.80\% & 22.69\% & 25.10\% \\
 & Early end \% & 30.14\% & 37.08\% & 38.56\% \\
 & Late end \%~ & 28.60\% & 24.08\% & 28.74\% \\ 
\hline
\multirow{3}{*}{Mean Absolute Error} & MAE in start time & 19.0 & 19.1 & 22.2 \\
 & MAE in end time & 21.3 & 22.4 & 25.6 \\
 & MAE in centre & 18.6 & 19.2 & 22.3
\end{tabular}
}
\caption{\label{tab:error-timestamp}
Error analysis of the predictions compared to the ground truth (GT). The top half of the table is the percentage of predictions starting/ending earlier/later than the GT. The margin is set to be 10\% of the video length to be classified as early or late for this analysis. 
The bottom half of this table reports the Mean Absolute Error (MAE) in the start/end/center times. 
All error analysis is performed for the three different variations of the vision-language connector.
}
\end{table*}

\begin{table}
\centering
%\resizebox{\linewidth}{!}{%
\begin{tabular}{c|cc}
Method & MSVD-QA & ActivityNet-QA \\ 
\hline
Video-LLaMA2\cite{videollama2} & 82 & 80 \\
Video-LLaVA\cite{videollava} & 81 & 74 \\
\textbf{Ours} & \textbf{88} & \textbf{83}
\end{tabular}
%}
\caption{\label{tab:llama-open-qa}
Comparison with other methods on the open-ended QA task, evaluated using Llama 3.1. The metric reported here is the QA accuracy as judged by Llama. We observe that Llama is more lenient than GPT in terms of evaluating responses. However, the trend is similar and our approach outperforms other methods on the open-ended video QA task.
}
\end{table}

\paragraph{Different downsampling rate}
We sample 16 frames at the dense scale, and 4 frames at the sparse scale - a (16, 4) configuration. Thus, we use 4X downsampling from dense to sparse.
In this section, we try a 2X downsampling rate - meaning we try a (16, 8) configuration. We observe that (16, 8) configuration achieves a mIoU of 21.7 on NExT-GQA, as compared to the mIoU of 24.4 achieved by the (16, 4) configuration.
We think that 2X downsampling is not sufficient for the sparse branch, making it difficult to model long-term temporal relations due to the large amount of information even at the sparse scale.
%\AG{Check this. Also, why do we not try 8X downsampling?}

\subsection{Novel open-Acc@GQA metric for NExT-GQA.}

Recall that to compute the Acc@GQA metric for NExT-GQA, the answer and the grounding both need to be evaluated simultaneously.
The grounding is deemed `correct' if the IoP $> 0.5$.
The answer is `correct' if the correct option is chosen among the five pre-determined options for the question.
An answer+grounding pair is correct for the Acc@GQA metric if both the two conditions are satisfied.

However, our approach generated free-form answers in an open-ended setup. To compute the Acc@GQA metric and compare our method to other methods, we have to `choose' an option after generating open-ended answers. Recall that we perform GPT-assisted retrieval for this, retrieving the most similar option to the open-ended answer generated by our method. 

In this section, we propose a new metric more suitable for evaluating the open-ended nature of our approach, called \textit{Open-Acc@GQA.}
This is based on the GPT-assisted open-ended QA evaluation performed in other works \cite{videollama2, videollava, videochatgpt}. Specifically, we query GPT to compare our response and the ground truth answer, given the question.
We ask GPT to come up with a `yes' or `no' response to whether the prediction is similar in meaning to the ground truth.
Note that, as opposed to the GPT-assisted retrieval performed earlier, the options are not passed to GPT. 

We impose similar conditions on the Open-Acc@GQA metric, as imposed by Acc@GQA. Specifically, the predictions are to be similar in meaning to the ground truth (as described earlier) and the IoP of the grounding must be $> 0.5$. In this way, Open-Acc@GQA becomes more suited to evaluate the open-ended grounded video QA task.
We achieve \textbf{21.4\%} on the Open-Acc@GQA metric. Note that we can not directly compare this to existing approaches for grounded video QA since they do not generate open-ended answers. 
It is interesting to compare this number to the Acc@GQA metric, on which the same model achieves 24.6\%.
Open-Acc@GQA is a more difficult metric than Acc@GQA due to the lack of options in the former.

% Compared to the Acc@GQA metric
% \AG{Compare to Acc@GQA 24.6, mention why less}

\subsection{Prediction statistics and error analysis}

We report statistics of the predicted and ground truth timestamps in \cref{tab:stats-gt-pred-timestamps}. 
We notice that the average length and mean center position are similar for both, but predictions are slightly biased toward the early start and late end.

We also perform some additional error analysis on the predicted timestamps to get insights into where the model is going wrong.  
We compute the percentage of times the model makes an early/late prediction for start/end times. 
We also compute the mean absolute error (MAE) in the start, end, and center of the predicted timestamp compared to the GT.
This analysis is presented in \cref{tab:error-timestamp}. These observations entail that errors are mostly uniform in either direction for both start and end times.

% \begin{table}
% \centering
% %\resizebox{\linewidth}{!}{%
% \begin{tabular}{c|c|c}
% \multirow{4}{*}{Prediction mismatch \%} & Early start ~ & 24.90\% \\
%  & Late start ~ & 26.80\% \\
%  & Early end ~ & 30.14\% \\
%  & Late end ~ & 28.60\% \\ 
% \hline
% \multirow{3}{*}{Mean absolute Error} & MAE in start time & 19.0 \\
%  & MAE in end time & 21.3 \\
%  & MAE in centre & 18.6
% \end{tabular}
% %}
% \caption{\label{tab:error-timestamp}
% Error analysis of the predictions compared to the ground truth (GT). The top half of the table is the percentage of predictions starting/end earlier/later than the GT. The margin is set to be 10\% of the video length to be classified as early or late for the purpose of this analysis. 
% The bottom half of this table reports the Mean Absolute Error (MAE) in the start/end/centre times. 
% }
% \end{table}

\subsection{Open-ended Llama-based evaluation}
\label{sec:llama-eval}
We have evaluated the open-ended video QA task using GPT-assisted evaluation, to be consistent with previous work \cite{videollama2, videollava, videochatgpt}. However, GPT is a closed source and is expensive to use. 
Thus, to improve accessibility and reproducibility, we also evaluate our approach using the open-source pretrained LLM Llama 3.1. Similar to the GPT evaluation, we pass the question, prediction, and ground truth to Llama. We ask it to come up with a `yes' or `no' response - to whether the prediction is similar to the ground truth answer. The percentage of `yes' responses is the accuracy of the model. 
To compare with other approaches for open-ended video QA, we also reproduce other approaches and evaluate them using the same Llama-based evaluation. 

As shown in \cref{tab:llama-open-qa} Llama is generally more lenient in evaluation compared to GPT, since the numbers for all methods are higher than the corresponding GPT-assisted evaluation.  Nonetheless, we observe a similar trend as the GPT evaluation and our approach outperforms others in this evaluation.

%\AG{Reason: GPT has 175B parameters, Llama has 7B?}

\subsection{Effect of chat template}
Recall that we apply a chat template to every question before passing it to the model. 
In the template, we include the desired output format, eg. \texttt{answer} or \texttt{answer [<start>, <end>]}. 
While evaluating on NExT-GQA, we generate the answers and the grounding jointly, so we include the latter format in the queries in NExT-GQA. 
In this section, we try an experiment where we do not apply any template to the question.
We observe that the model fails to predict the grounding for any query in the NExT-GQA test set.
Thus, we conclude that the template is essential for the model to generate temporal grounding along with the answers. 

\subsection{Acc@QA metric}
The Acc@GQA metric evaluates the grounding and QA performance simultaneously. However, another metric Acc@QA is also used in \cite{nextgqa-dataset} to evaluate just the QA ability separately. For this, we discard the temporal groundings in the generated answer and just evaluate the answers using our GPT-assisted retrieval method. We achieve an accuracy of 67.0\%, which is close to the SOTA \cite{llovi} which achieves 67.7\%. 
Note that our model does not consider options while generating the answer, as opposed to \cite{llovi}. 

\subsection{Frame numbers in prompt}
\label{sec:interleave-expt}
Taking inspiration from \cite{mr-blip-grounding-llm, lita}, we try to input the frame indices between the visual features in the prompt, to see if it further improves the grounding performance. 
However, we observe that this decreases performance - achieving a mIoU/mIoP of 21.2/36.4 respectively, as opposed to 24.4/40.5 of our best model. 
We observe that these approaches work with image-level features, and interleave the frame indices between features of individual frames.
However, our multi-scale connector comprises of 3D convolutions that learn the correlations between frame features within the connector itself. 
Hence, explicit frame indices are less important in the prompt, since the temporal information is already present in 3D convolutional features.

\subsection{Effect of language decoder}
% \AG{Text decoder (vicuna v1.3) ablation - 19.0 miou and 33.9 mIoP}
We have used the Mistral-7B \cite{mistral-7b} language decoder in our experiments. We also experiment with another popular decoder, namely Vicuna-7B. We find that using the vicuna-7B language decoder in our approach achieves a mIoU/mIoP of 19.0/33.9 respectively, which is lower than our main model (24.4/40.5). 

\subsection{Traditional metrics for QA evaluation}
Most of the Video QA works use GPT for evaluation \cite{videollama2, videollava}. However, it may be seen as inconsistent over time due to API version changes in the closed-source GPT model. Hence, we also evaluate using open-source LLaMA in \cref{sec:llama-eval}. 
Further, we utilize the traditional METEOR metric \cite{meteor} for a more robust evaluation of our model. We achieve a METEOR score of 0.498 on the ActivityNet-QA dataset.

\subsection{Evaluation on additional grounding dataset}
We also evaluate our approach on the ReXTime \cite{rextime} dataset in a zero-shot setting, without finetuning on these videos. We get a mIoU of 27.4 and mIoP of 41.9 on the validation set of ReXTime without finetuning. 
We were not able to evaluate on the test set through the submission website, hence we evaluated on the validation set results. 
% \AG{Mention no other work has evaluated on test set of this benchmark? Is this okay?}

\section{Qualitative examples}
% \AG{More activitynet examples from slides - with grounding?}

\begin{figure*}[t]
    \centering
    % Top Image
    \includegraphics[width=0.7\linewidth]{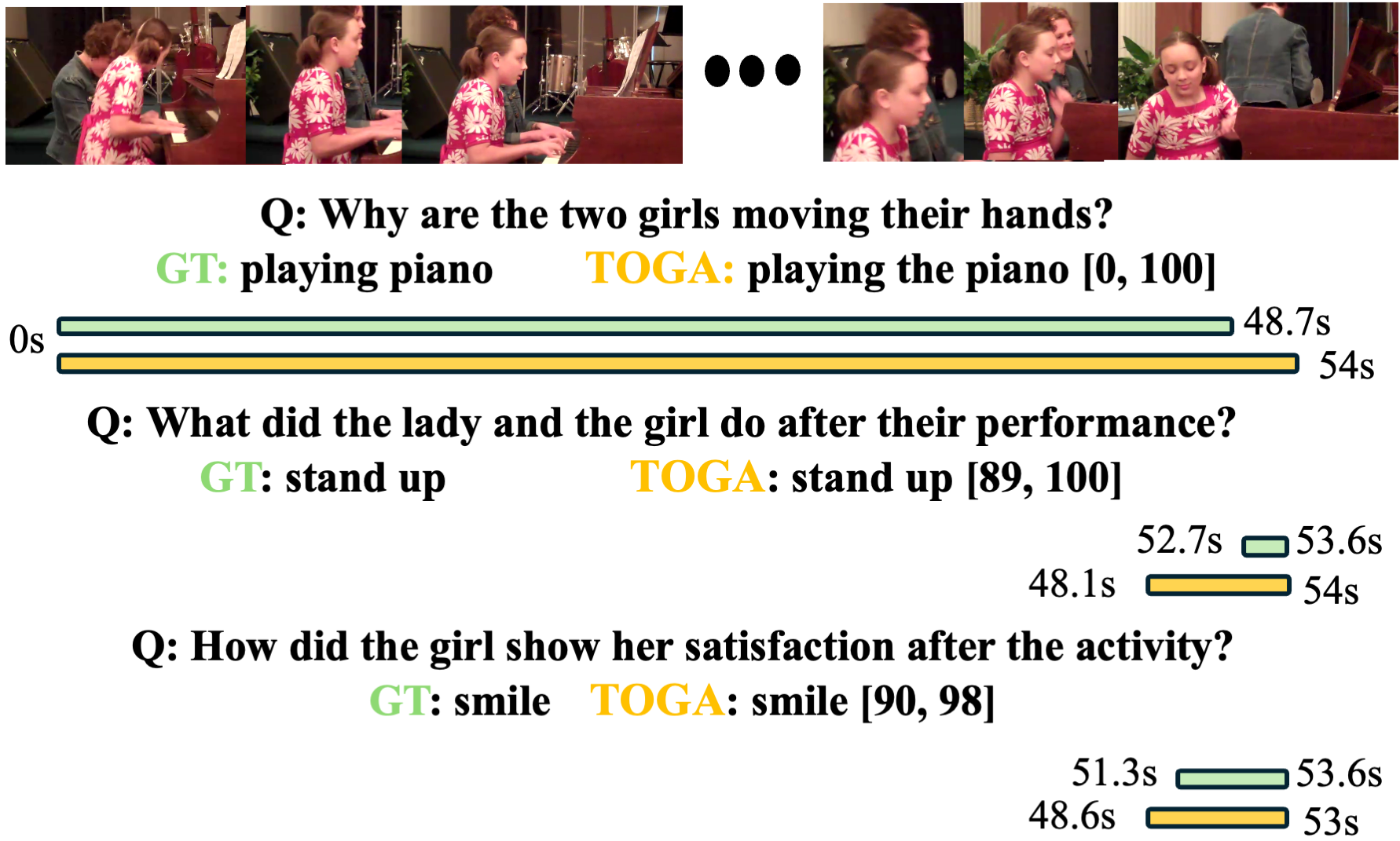}
    \vspace{0.5cm} % Space between images
    % Bottom Image
    \includegraphics[width=0.7\linewidth]{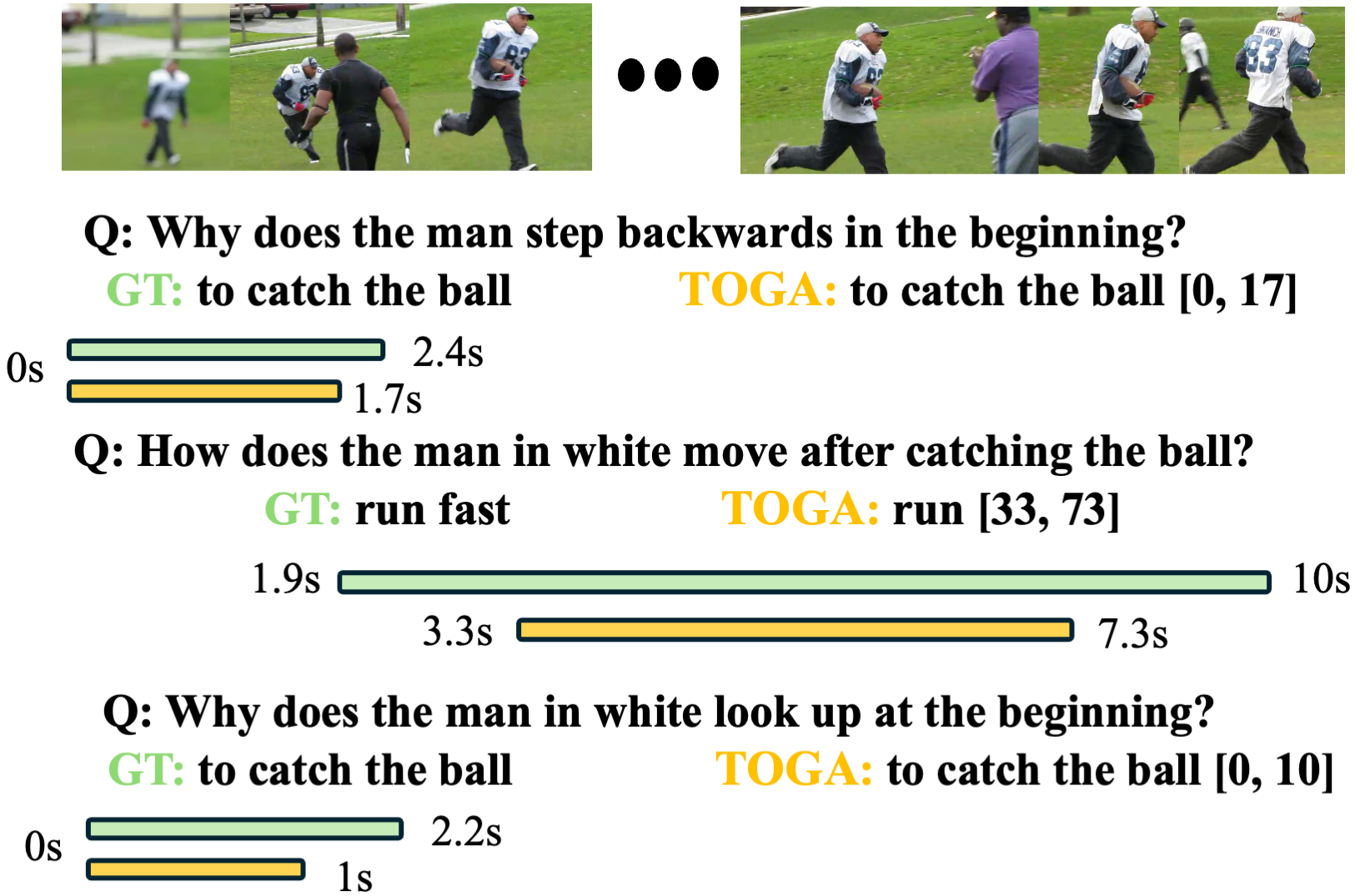}
    \caption{Additional qualitative examples from the NExT-GQA dataset, with longer videos and multiple questions per video. Compared to the length of the video, the grounding evidence for the answer may be small. However, our model can capture events in small temporal windows as well.}
    \label{fig:iccv-qual-supp-nextgqa}
\end{figure*}

\begin{figure*}
    \centering
    \includegraphics[width=0.9\linewidth]{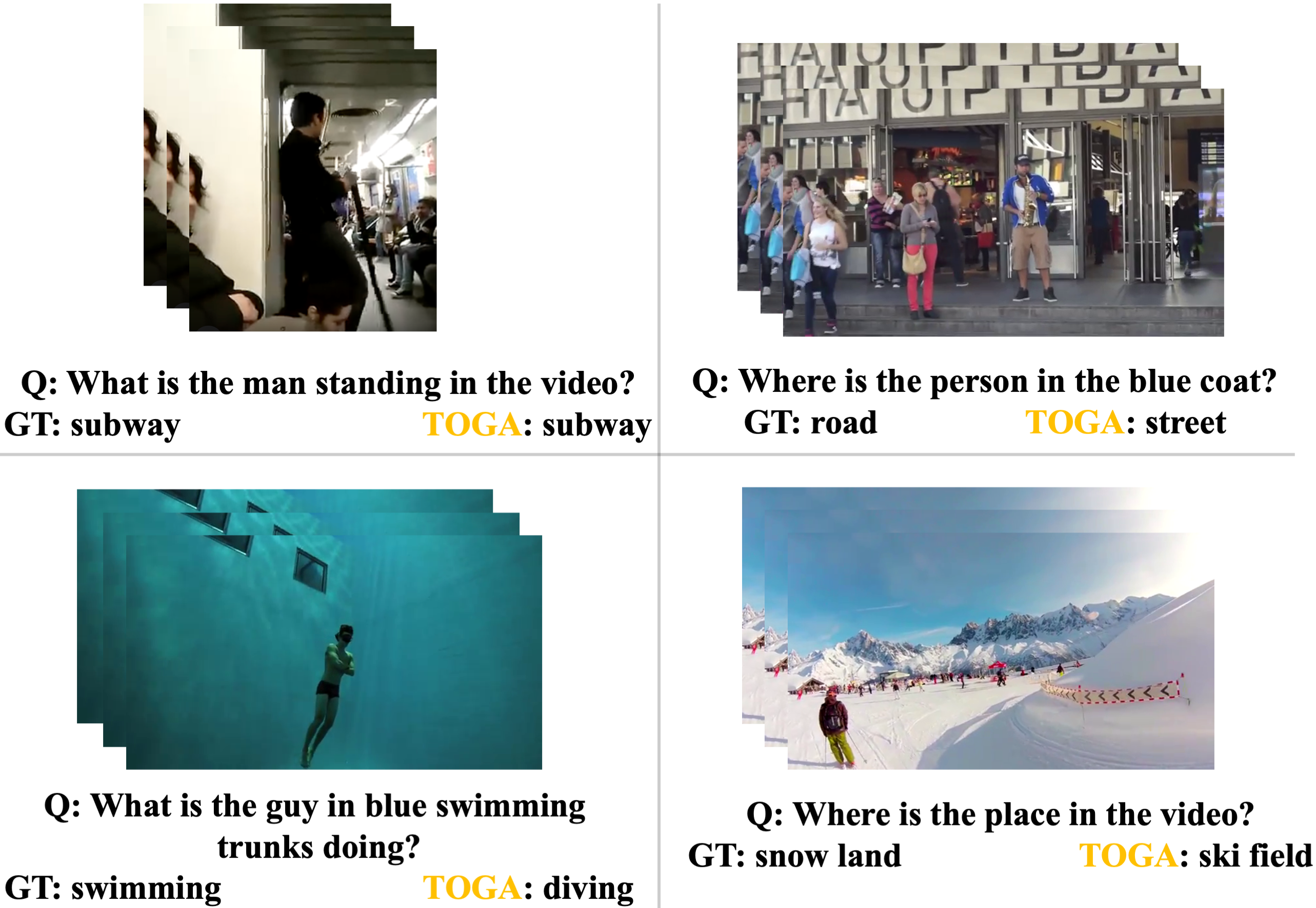}
    \caption{Additional qualitative examples from the ActivityNet dataset. Our model may generate words that are similar in meaning but not the same as the ground truth, due to the open-ended nature of our approach.}
    \label{fig:qual-supp-actnet}
\end{figure*}

We include additional qualitative examples for the NExT-GQA dataset in \cref{fig:iccv-qual-supp-nextgqa} and the ActivityNet dataset in \cref{fig:qual-supp-actnet}. 
We include three questions for each video in \cref{fig:iccv-qual-supp-nextgqa}. 
We can observe that the model can detect events that happen in a relatively short temporal window compared to the length of the video. 
We can also observe that the predicted answers may not exactly match the ground truth in both these figures. This is due to the open-ended nature of our approach. 

\begin{figure*}
    \centering
    \resizebox{0.8\linewidth}{!}{
    \includegraphics{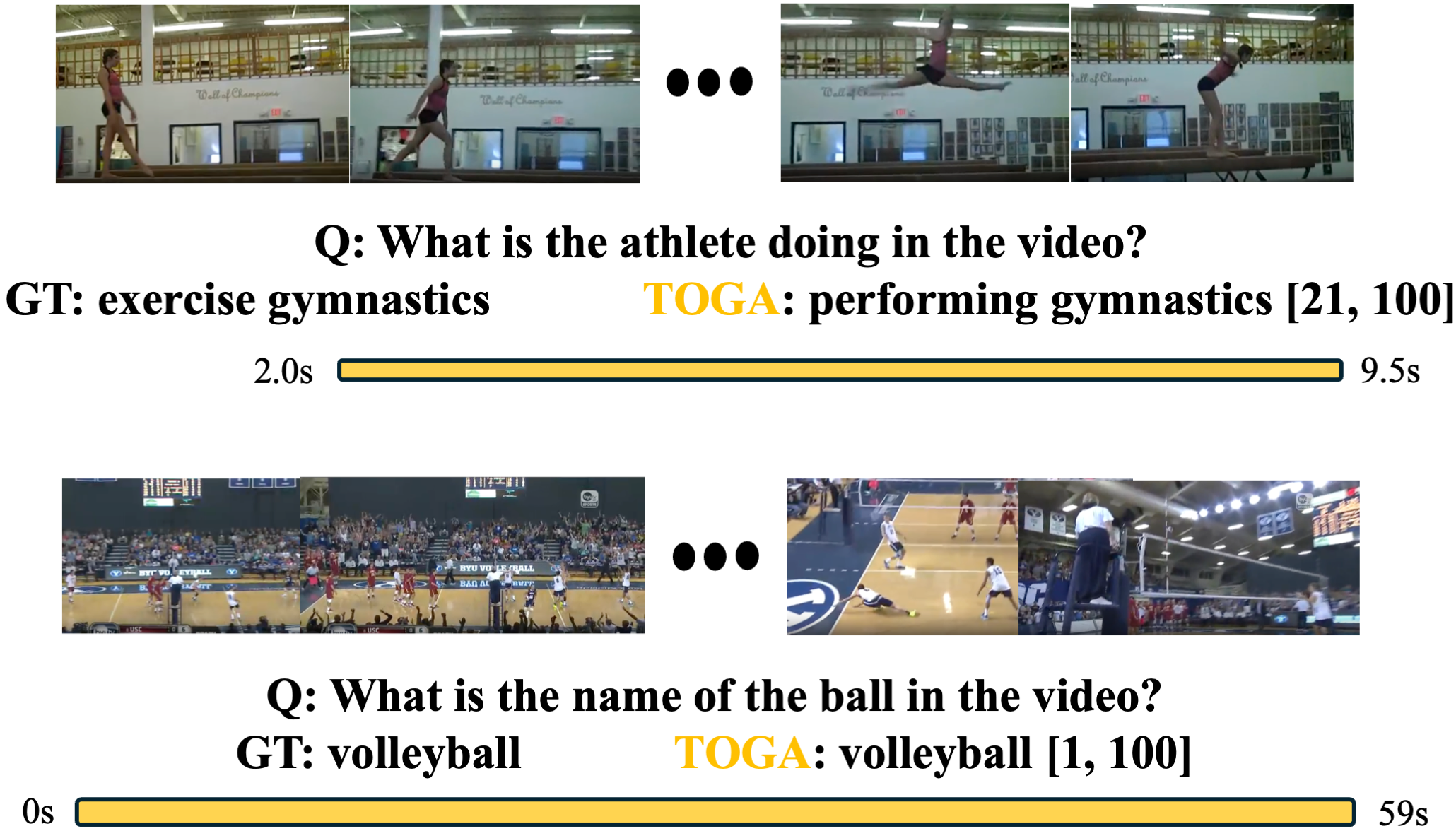}
    }
    \caption{Examples of grounding on ActivityNet videos. \approach can perform grounding even on this dataset, but the questions may be relevant to the whole video rather than a part of it.}
    \label{fig:actnet-grounding-ex}
\end{figure*}

Even though ActivityNet is not a grounding dataset, we still try to ground the questions in the dataset to observe the performance. We show examples in \cref{fig:actnet-grounding-ex}. \approach generates some groundings, but we can not evaluate the quality without the ground truth annotations. Manually looking at these samples, the predictions seem reasonable based on the correlation between answers and the frames.

% \begin{figure*}
%     \centering
%     \includegraphics[width=\linewidth]{figures/supp-nextgqa-ex.png}
%     \caption{Additional qualitative examples from the NExT-GQA dataset, with longer videos and multiple questions per video. Compared to the length of the video, the grounding evidence for the answer may be very small. But our model is able to capture events in small temporal windows as well.
%     \AG{Restructure, move to end}
%     }
%     \label{fig:qual-supp-nextgqa}
% \end{figure*}

\end{document}